\pdfoutput=1
\documentclass[11pt]{article}

\usepackage{acl}

\usepackage{times}
\usepackage{latexsym}

\usepackage[T1]{fontenc}
\usepackage[utf8]{inputenc}

\usepackage{microtype}

\usepackage{inconsolata}
\usepackage{todonotes}
\usepackage{rotating}
\usepackage{graphicx}
\usepackage{balance} % for balancing columns on the final page
\usepackage{lscape}
\usepackage{amsmath}
\usepackage{amssymb}
\usepackage{amsthm}
\usepackage{float}
\usepackage{multirow}
\usepackage{algorithm}
\usepackage{algpseudocode}
\usepackage{tcolorbox}
\usepackage{booktabs}
\usepackage{pifont}
\usepackage{bbm}
\usepackage[normalem]{ulem}
\usepackage{hyperref}
\usepackage{algorithmicx}
\useunder{\uline}{\ul}{}

\newcommand{\KG}{\ensuremath{\mathcal{G}}}

\algdef{SE}[DOWHILE]{Do}{doWhile}{\algorithmicdo}[1]{\algorithmicwhile\ #1}%

\newtheorem{theorem}{Theorem}

\newtheorem{corollary}{Corollary}
\algnewcommand{\LineComment}[1]{\State \(\triangleright\) #1}
\title{ Conformalized Answer Set Prediction for Knowledge Graph Embedding }

\author{%
  Yuqicheng Zhu$^{\dagger\ddagger}$, Nico Potyka$^\mathsection$, Jiarong Pan$^{\ddagger *}$,  Bo Xiong$^\diamond$ \\
   \textbf{Yunjie He}$^{\dagger\ddagger}$\textbf{,} \textbf{Evgeny Kharlamov}$^{\ddagger\flat}$\textbf{,} \textbf{Steffen Staab}$^{\dagger\natural}$
   \\
   $^\dagger$University of Stuttgart, $^\ddagger$Bosch Center for AI, $^\mathsection$Cardiff University, $^\flat$ University of Oslo, \\$^\natural$ University of Southampton, $^*$ Eindhoven University, $\diamond$ Stanford University\\
  \texttt{yuqicheng.zhu@de.bosch.com}\\
}

\begin{document}
\maketitle
\begin{abstract}
Knowledge graph embeddings (KGE) apply machine learning methods on knowledge graphs (KGs) to provide non-classical reasoning capabilities based on similarities and analogies. 
%\todo[inline]{based on reviews, perhaps remove the following sentence because our approach does not help in giving a meaningful ranking}
The learned KG embeddings are typically used to answer queries by ranking all potential answers,
but rankings often lack a meaningful probabilistic interpretation - lower-ranked answers do not necessarily have a lower probability of being true.
%but the ranking is typically not meaningful
%in the sense that a lower-ranked answer does not necessarily have lower probability of being true. 
This limitation makes it difficult to quantify uncertainty of model's predictions, posing challenges for the application of KGE methods in high-stakes domains like medicine.
%We instead quantify the uncertainty of the KGE models' predictions by 
%guarantee a higher correctness probability.\todo{remember to improve}
% which is not always practical for decision-making that requires reliable answer sets. 
%To construct answer sets with formal guarantees, we 
We address this issue by applying the theory of conformal prediction that allows generating answer sets, which contain
the correct answer with probabilistic guarantees.
We explain how conformal prediction can be used to generate such answer sets for link prediction tasks. 
% These sets reliably contain the true answer with a user-specified confidence level, such as $90\%$. 
Our empirical evaluation on four benchmark datasets using six representative KGE methods validates that the generated answer 
sets satisfy the probabilistic guarantees given by the theory of conformal prediction.
We also demonstrate that the generated answer sets often have a sensible size and that the size adapts well with
respect to the difficulty of the query.
% our approach consistently 
% meets the probabilistic guarantees of conformal prediction. Moreover, it provides high-quality prediction sets, evaluated in terms of their average size and adaptiveness, outperforming other baselines. 
\end{abstract}

\section{Introduction}
%1 What is the problem?
Knowledge Graph Embeddings (KGE) map entities and predicates into numerical vectors, providing non-classical reasoning capabilities by exploiting similarities and analogies between entities and relations \citep{wang2017knowledge, biswas2023knowledge}. 
KGE models are typically evaluated through link prediction \cite{bordes2013translating, sun2019rotate, nickel2011rescal}. To answer queries in the form of $\langle \textit{head entity}, \textit{predicate}, ?\rangle$ or $\langle ?, \textit{predicate}, \textit{tail entity}\rangle$, all possible entities are ranked according to their plausibility scores returned by the KGE models. 
% \todo[inline]{is the following sentence comprehensible? What does model performance mean here?}
% Higher-ranked positions of the correct answers indicate better model performance.
The higher the correct answer is ranked, the more precise the model's prediction is for that particular query.

%2 Why is it interesting and important?
However, high-ranked entities does not necessarily correlate with high likelihood of correctness \cite{Tabacof2020calibration, Safavi2020calibration}. 
For instance, in a medical diagnosis setting, ranking "common cold" above "cancer X" may imply various scenarios:
(1) the common cold is highly likely while cancer is improbable, 
(2) both conditions are likely but the common cold is more probable, 
or (3) neither is likely, but the common cold is relatively more probable. 
These interpretations can lead to vastly different treatments, making rankings alone insufficient for medical decisions. It is especially important for the patient to know when cancer can be confidently excluded from consideration.

% \todo[inline]{Is the following really a good motivation? Can the entities in our prediction sets be called plausible or implausible?}
% \todo[inline]{If we have a single label/number as prediction, it is easy to motivate - we need to quantify uncertainty of the single estimate, namely a prediction interval, however, I can not connect with rankings. }
% However, ranking all possible entities has limited practical value, since the rankings do not distinguish plausible answers from implausible ones with high quality. 
% \todo[inline]{The following sounds convincing, but is not really connected to the former. Perhaps just start with this idea? }
%It is crucial to provide tight answer sets that provably cover the true answer
%, particularly in high-stakes domains like medicine, where reliable predictions and risk assessment are critical. 

%\todo[inline]{How does our approach help with the following scenario? Couldn't common cold and cancer X also occur together in a prediction set?}

%\todo[inline]{Perhaps remove the calibration paragraph or rewrite? }
%3 	Why is it hard? (E.g., why do naive approaches fail?)
Although uncertainty quantification is crucial for high-stakes applications like in medicine domain, where reliable predictions and risk assessment are essential, methods for quantifying uncertainty in KGE model predictions remain underexplored.
Existing approaches often reply on off-the-shelf calibration techniques, such as Platt scaling \cite{platt1999probabilistic} and Isotonic regression \cite{kruskal1964nonmetric}, to map uncalibrated plausibility scores to the expected correctness of predictions \cite{Tabacof2020calibration, Safavi2020calibration}. However, perfect calibration is impossible in practice \cite{gupta2020distribution}, both Platt scaling and Isotonic regression are empirical calibration techniques that lack formal probabilistic guarantees and are highly sensitive to the calibration set. 
%Identifying a set of plausible answer entities from the entire entity set is challenging. 
% KGE models are trained to assign higher plausibility scores to true triples than to false ones. 
% However, due to the lack of ground truth negative triples in KGs, negative examples used for training are typically generated by corrupting existing triples \cite{bordes2013translating, sun2019rotate}. Some of the generated negative samples may actually be valid but unobserved triples, which can mislead the KGE model and result in incorrect plausibility scores.
% Furthermore, the training process uses gradient descent-based optimization techniques \cite{rumelhart1986learning}, which do not guarantee convergence to the global optimum.
% As a result, the plausibility scores returned by KGE models not only fail to ensure correct triple ranking, but they also lack a clear probabilistic interpretation, meaning they do not correspond to the actual likelihood of a triple being true (i.e., they are uncalibrated).
%Since the plausibility scores returned by KGE models lack a clear probabilistic interpretation, meaning they do not correspond to the actual likelihood of a triple being true (i.e., they are uncalibrated) \cite{Tabacof2020calibration, Safavi2020calibration}.
%Therefore, constructing answer sets based on the uncalibrated plausibility scores can lead to somewhat arbitrary result sets.
%4 Why hasn't it been solved before? (Or, what's wrong with previous proposed solutions? How does mine differ?) 

%5 What are the key components of my approach and results? Also include any specific limitations.
In contrast, our work diverges from these approaches by applying the theory of conformal prediction \cite{vovk2005algorithmic} to quantify uncertainty with formal statistical guarantees. 
Conformal prediction assigns a score to each candidate answer entity and defines a threshold to construct answer sets that include the truth answer with a desired confidence level.
The size of the answer set reflects the uncertainty of the model's output for a given query.
%the answer set contains the answers that conforms most to the existing query-answer pairs (triples) in the training set. The degree of conformity is adjusted by a user-specified error rate $\epsilon$. Through conformal prediction, it is guaranteed that the answer set contain the true answer with a probability of at least $1-\epsilon$.
%conformal prediction tests whether each possible answer to a query conforms to the existing query-answer pairs (triples) in the training set. The answer set is then constructed by including all answers that do not reject the true hypothesis. 
To the best of our knowledge, this is the first method that does not merely convert plausibility scores into probabilities but instead ensures statistical validity in the uncertainty quantification of the predictions within the context of KGE.

%Conformal prediction is a general framework rather than a specific algorithm. 
In our paper, we carefully design conformal predictors tailored to the link prediction task such that the answer sets are (1) probabilistically guaranteed to include the true answer entity at a specified confidence level, (2) tight, and (3) adaptive, providing smaller sets for easier queries than harder ones. We perform extensive experiments on commonly used benchmark datasets and a variety of KGE methods. Our empirical results show that: (1) conformal predictors satisfy the statistical guarantees in Proposition \ref{prop:cover} and produce tighter answer sets compared to other baselines (Experiments 1); (2) conformal predictors generate answer sets that adapt to query difficulty, yielding smaller sets for easier queries than for harder ones (Experiment 2); (3) high-quality answer sets can be obtained with a relatively small calibration set (Experiment 3); and (4) conformal predictors are effective across different user-specified error rates (Experiment 4).

\section{Related Work}
As highlighted in recent work \cite{zhu2024predictive}, predictions from KGE models exhibit substantial uncertainty, even among models with similar overall performance. This phenomenon, termed predictive multiplicity, arises from the highly non-convex loss functions employed by KGE methods. It leads to multiple models that capture distinct patterns from the training KG and generalize differently. Consequently, quantifying the uncertainty of KGE methods is critical; however, this remains a relatively underexplored area of research.

Existing approaches incorporate uncertainty into KGE by modeling entities and relations using probability distributions \cite{he2015learning, xiao2015transg}. However, these methods primarily focus on enhancing the performance of KGE models through more expressive representations, without systematically analyzing or rigorously evaluating the quality of uncertainty in embeddings or predictions.

Furthermore, research by \citet{Tabacof2020calibration} and \citet{Safavi2020calibration} applies off-the-shelf calibration techniques, such as Platt scaling and Isotonic regression, to KGE methods. These techniques aim to convert uncalibrated plausibility scores into probabilities by minimizing the negative log-likelihood on a validation set. However, these approaches are quite sensitive to the validation set and do not provide formal guarantees about the generated probabilities.

This paper applies conformal prediction, which has its roots in online learning literature, is a method that produces predictive sets ensuring coverage guarantees \cite{vovk2005algorithmic}. 
This approach has been successfully applied across various domains, including image classification \cite{Angelopoulos2021image}, natural language processing \cite{maltoudoglou2020bert} and node classification/regression on graphs \cite{huang2024uncertainty, zargarbashi2023conformal, zargarbashi2023inductive}. However, to the best of our knowledge, it has not yet been applied to KGE.

\section{Preliminaries}\label{sec:pre}

\subsection{Knowledge Graph Embedding}
We consider a knowledge graph (KG) $\KG \subseteq E \times R \times E$ defined over a set $E$
of entities and a set $R$ of relation names. The elements in $\KG$ are 
called triples and denoted as $<h,r,t>$.
A KGE model $M_\theta:E\times R\times E\rightarrow\mathbb{R}$ associates each triple with a 
%\todo{termniology: in this context, it may be confusing to refer to the output of the KGE model as a predictive score. Perhaps 'plausibility' for the KGE model and 'predictive' for conformal prediction?} 
score that measures the plausibility that the triple holds. 
The parameters $\theta$ are learned to let $M_\theta$ assign higher plausibility scores to positive triples (real facts) while assigning lower plausibility scores to negative triples (false facts).

Note that the interpretation of plausibility scores varies across different types of KGE methods. In \emph{distance-based models} like TransE \cite{bordes2013translating} and RotatE \cite{sun2019rotate}, the plausibility score is determined by the negative distance in the embedding space. In \emph{semantic matching models} such as RESCAL \cite{nickel2011rescal} and DistMult \cite{yang2015distmult}, plausibility scores are derived from similarity measures. %(often computed through the dot product of entity embeddings).

%\todo[inline]{explain the assumption of the output, already normalized? what are the output?}

% This can be achieved for example by minimizing the \emph{margin-based ranking loss} \cite{bordes2013translating}:
% \begin{equation}
%     \mathcal{L}=\sum_{tr\in\postriple}\sum_{tr^-\in\negtriple}\max(0,\gamma-M_\theta(tr)+M_\theta(tr^-)),
% \end{equation}
% or the \emph{cross-entropy loss} \cite{trouillon2016complex}:
% \begin{equation}
%     \mathcal{L}=\sum_{tr\in\postriple\cup\negtriple}\log(1+\exp(-y_{tr}\cdot M_\theta(tr))),
% \end{equation}
% where $\gamma$ is a margin hyperparameter, $tr$ refers to a triple $\langle h,r,t\rangle$, $\postriple, \negtriple$ are the sets of positive and negative triples, respectively. The label of a triple, denoted as $y_{tr}$, takes values from the set $\{-1, 1\}$. Here, $y_{tr} = 1$ indicates the triple as positive, while $y_{tr} = -1$ indicates that the triple is negative. The negative triples are typically generated by randomly replacing the head entity or the tail entity in a positive triple with a random entity sampled from the entity set.

\subsection{Conformal Prediction}\label{sec:cp}
Conformal prediction (a.k.a conformal inference) is a general framework for producing answer sets that cover the ground truth with probabilistic guarantees \cite{vovk2005algorithmic}. In this section, we recall some basics from \cite{vovk2005algorithmic, shafer2008tutorial}.

Let $(x_i, y_i)$ denote a data point with an object $x_i$ and its label $y_i$. The objects are elements of an \emph{object space} $\mathcal{X}$, and the labels are elements of a \emph{label space} $\mathcal{Y}$. 
For a more compact notation, we write $z_i$ for $(x_i, y_i)$, and call $\mathcal{Z}:=\mathcal{X}\times\mathcal{Y}$ the \emph{example space}. Furthermore, we let $Z_{1:n}=\{z_1,\dots z_n\}\subseteq\mathcal{Z}$ be the set of $n$ examples 
and denote $\mathcal{Z}^*$ as the set of all possible example sets. 

% Formally, a conformal predictor $\Gamma$ is defined as a measurable function:
% \begin{equation}
%     \Gamma: \mathcal{Z^*}\times\mathcal{X}\rightarrow 2^{\mathcal{Y}},
% \end{equation}
% where $2^{\mathcal{Y}}$ is the set of all subsets of $\mathcal{Y}$. 
A conformal predictor $\Gamma: \mathcal{Z^*}\times\mathcal{X}\rightarrow 2^{\mathcal{Y}}$ aims to predict a subset of $\mathcal{Y}$ large enough to cover the ground truth with high probability. 
Given a training set $Z_{1:n}$ and any new object $x_{n+1}\in\mathcal{X}$, 
%For every probability of error $\epsilon\in (0,1)$, $\Gamma^\epsilon$ is a $(1-\epsilon)$-prediction set; it contains
% Given a training set $Z_{1:n}$, a test object $x_{n+1}$ and a significance level $\epsilon\in (0,1)$, $\Gamma$ predicts that\todo{this sentence is hard to understand (we can talk about it)}
% \begin{equation}
%     y_{n+1}\in\Gamma^\epsilon(Z_{1:n}, x_{n+1}),
% \end{equation}
% the smaller $\epsilon$ is the more confident the prediction set will cover the true label $y_{n+1}$. 
% The prediction sets are required to shrink as $\epsilon$ increases: $C^\epsilon$ must satisfy 
% \begin{equation}
%     C^{\epsilon_1}(x_{n+1})\subseteq C^{\epsilon_2}(x_{n+1}), 
% \end{equation}
% whenever $\epsilon_1\geq\epsilon_2$.
the conformal predictor $\Gamma$ should, for every probability of error $\epsilon\in (0,1)$, produce a answer set $\Gamma^\epsilon(Z_{1:n}, x_{n+1})$ for the input object $x_{n+1}$ that contains the ground truth label $y_{n+1}$ with probability at least $1-\epsilon$. Moreover, the answer sets are required to shrink as $\epsilon$ increases: $\Gamma^{\epsilon_1}\subseteq\Gamma^{\epsilon_2}$ whenever $\epsilon_1\geq\epsilon_2$.

%The conformal predictor $\Gamma$ produces a prediction set $\Gamma^\epsilon$ for every probability of error $\epsilon\in (0,1)$, which contain the ground truth with probability at least $1-\epsilon$. Moreover, the prediction sets are required to shrink as $\epsilon$ increases: $\Gamma^{\epsilon_1}\subseteq\Gamma^{\epsilon_2}$ whenever $\epsilon_1\geq\epsilon_2$.

To specify such a conformal predictor, we first need to define a \emph{nonconformity measure} $S:\mathcal{Z}^*\times\mathcal{Z}\rightarrow\overline{\mathbb{R}}$.
$S(Z_{1:n}, z_{n+1})$ measures how unusual the example $z_{n+1}$ is as an element of $Z_{1:n}$. 
% For example, we can measure the nonconformity of the new example $z_{n+1}$ to the training set $Z_{1:n}$ by comparing the distance of object $x_{n+1}$ to training objects with the same label to its distance to those with a different label:
% \begin{equation}
%     \small S(Z_{1:n}, z_{n+1}) = \frac{\min\{|x_i-x_{n+1}|: 1\leq i\leq n, y_i=y_{n+1}\}}{\min\{|x_i-x_{n+1}|: 1\leq i\leq n, y_i\not=y_{n+1}\}}.
% \end{equation}
%
% \todo[inline]{Essentially, conformal prediction calculate the nonconformity scores of all examples in $Z_{1:n}$, when testing new example, we calculate the nonconformity score of the example, conduct hypothesis testing to decide whether to include the label in prediction set.}
% Given a nonconformity measure $S$ and a set of triples $Z_{1:n}$, the nonconformity score of $z_i\in \mathcal{Z}$ is defined as
% \begin{equation}
%     \alpha_i := S(Z_{1:n}, z_i).
% \end{equation}
%
% The fraction of the examples in $Z_{1:n}$ as nonconforming as $z_i$
% \begin{equation}
%     \frac{|\{j=1,\dots,n:\alpha_j\geq\alpha_i\}|}{n}
% \end{equation}
% is called \emph{p-value} for $z_i$. The p-value of $z_i$ is small if $z_i$ is very nonconforming (an outlier). 
Given any such a nonconformity measure $S$, if we construct the answer set $\Gamma^\epsilon(Z_{1:n}, x_{n+1})$ by including all $y\in\mathcal{Y}$ such that 
\begin{equation}\label{eq:construct}
    \frac{|\{i=1,\dots,n+1:\alpha_i\geq\alpha_{n+1}\}|}{n+1}>\epsilon,
\end{equation}
where
\begin{align*}
    &\alpha_i:=S(Z_{1:n}\cup \{(x_{n+1}, y)\}, (x_i, y_i)), i=1,\dots,n\\
    &\alpha_{n+1}:=S(Z_{1:n}\cup \{(x_{n+1}, y)\}, (x_{n+1}, y)),
\end{align*}
then we have following probabilistic guarantees:
\begin{theorem}[\citet{vovk2005algorithmic, lei2018distribution}]\label{th:coverage}
    Suppose $n$ is large, and a set of examples $Z_{1:n+1}$ are independent and identically distributed (i.i.d.). Given $\epsilon\in(0,1)$, the answer set of the object $x_{n+1}$ constructed by a conformal predictor $\Gamma^{\epsilon}(Z_{1:n}, x_{n+1})$ cover the ground truth $y_{n+1}$ with a probability of at least $1-\epsilon$
    \begin{equation}\label{full_coverage}
        \mathbb{P}(y_{n+1}\in\Gamma^{\epsilon}(Z_{1:n}, x_{n+1}))\geq 1-\epsilon
    \end{equation}
    furthermore, if there are no ties between $\alpha_i$, then it is also holds that
    \begin{equation}\label{full_upperbound}
        \mathbb{P}(y_{n+1}\in\Gamma^{\epsilon}(Z_{1:n}, x_{n+1}))\leq 1-\epsilon+\frac{1}{n+1}
    \end{equation}
\end{theorem}
The proof of Equation \ref{full_coverage} is provided in \cite[section 2.1.3]{vovk2005algorithmic}. 
%\todo[inline]{Does the following really help the reader?}
Intuitively, the construction of $\Gamma^\epsilon(Z_{1:n}, x_{n+1})$ can be understood as an application of the widely accepted Neyman-Pearson theory \cite{lehmann1986testing} for hypothesis testing and confidence intervals \cite{shafer2008tutorial}. Here, we test for all $y\in\mathcal{Y}$ that the hypothesis $H$ (the example $(x_{n+1},y)$ conforms to $Z_{1:n}$) by evaluating the nonconformity score of $(x_{n+1},y)$. 
We construct the answer set by including all $y$, for which $(x_{n+1},y)$ is not rejected by the test. 

Additionally, the proof of Equation \ref{full_upperbound} is detailed in \cite[Appendix A.1]{lei2018distribution}. Notably, the theorem remains valid under the weaker assumption of \emph{exchangeability} \cite[section 2.1.1]{vovk2005algorithmic}. 

\section{KGE-based Answer Set Prediction}\label{sec:baselines}
%\todo{it may help the reader if you connect the earlier (x,y) notation to the triple notation explicitly}
In this section, we formally define the KGE-based answer set prediction task and outline three key desiderata guiding the development of effective set predictors. We then introduce and discuss several basic set predictors.
\subsection{Problem Definition and Desiderata}
We reformulate the link prediction task as an answer set prediction task. Instead of object-label pairs $(x_i,y_i)$ in section \ref{sec:pre}, each data point is a triple $tr(q_i,e_i)$. Here, $q_i$ denotes a query in form of either $\langle h,r,?\rangle$ or $\langle ?,r,t\rangle$, and $tr(q,e)$ corresponds to the respective triple $\langle h,r,e\rangle$ or $\langle e,r,t\rangle$.

Given a set of (training) triples $\mathcal{T}_{1:n}=\{tr(q_1,e_1), \dots, tr(q_n,e_n)\}$, a test query $q_{n+1}$ and a user-specific error rate $\epsilon$, we aim to predict a set of entities $\hat C(q_{n+1})\subseteq E$ that covers the true answer $e_{n+1}$ with probability at least $1-\epsilon$. 
\begin{equation}\label{eq:kge_cover}
    \mathbb{P}(e_{n+1}\in\hat C(q_{n+1}))\geq 1-\epsilon
\end{equation}

We refer to Equation \ref{eq:kge_cover} as the \emph{\textbf{coverage desideratum}}. However, this criterion alone is insufficient, as it can be trivially met by a predictor that always outputs sets containing all possible answers. To develop sensible set predictors, we also consider the \emph{\textbf{size desideratum}} and the \emph{\textbf{adaptiveness desideratum}}. The size desideratum emphasizes the need for smaller sets, as smaller sets are generally more informative. The adaptiveness desideratum requires that the set sizes reflect query difficulty: smaller sets should correspond to easier queries, while larger sets should be used for harder queries.

% \todo{shouldn't the baselines be discussed in the experiments after introducing your methods in the technical part?}
% \textbf{Random Baseline}. \todo{not sure, but are we confusing "the probability that a set contains the true label" with "the probability that a set that a method returns contains the true label" here?}This goal can be trivially achieved by predicting sets with all possible entities for $90\%$ queries and outputting random sets (to deal with the randomness, we always output empty sets in our implementation) for the rest. In this way, the model will cover the ground truth at least with probability $90\%$. Therefore, besides the coverage requirement in equation \ref{eq:kge_cover}, we also aim to reduce the size of the prediction sets. 
% \todo[inline]{remove this baseline}
\subsection{Basic Set Predictors}
\textbf{Naive Predictor}. Given a query, assume the KGE model provides the probability of each possible answer entity being true. A straightforward approach towards our goal is to construct the set by including entities from highest to lowest probability until their sum exceeds the threshold $1-\epsilon$.  
We refer to this approach as the \emph{naive} predictor and provide its pseudocode in Algorithm \ref{algo:naive} (Appendix \ref{app:code}).
However, the plausibility scores provided by KGE models are not calibrated. We convert these plausibility scores into a "probability distribution" using a softmax function. 

\textbf{Platt Predictor}. Following the recommendations of \cite{Tabacof2020calibration, Safavi2020calibration}, we improve the naive predictor by using a multiclass Platt scaling \cite{guo2017calibration} to calibrate the plausibility scores and then construct sets based on these calibrated probabilities. We refer to this method as the \emph{Platt} predictor and provide more details of this predictor in Appendix \ref{app:platt}.
%we enhance the naive predictor by applying a multiclass variant of Platt scaling \cite{guo2017calibration} to calibrate the plausibility scores of KGE models. We then construct the prediction sets based on calibrated probabilities. we refer to this method as the \emph{Platt} predictor and provide more details of this predictor in Appendix \ref{app:platt}.

\textbf{TopK Predictor}. Another straightforward approach is to construct the set with the Top-K entities from the ranking, referred to as the \emph{topk} predictor. We select $K$ to ensure the Top-K entities cover the correct answers for $1-\epsilon$ of the validation queries. %While this method may achieve the coverage desideratum, fixed-size prediction sets fail to reflect query-specific uncertainty, thus missing the adaptiveness desideratum.

\section{Conformal Prediction for KGE-based Answer Set Prediction}\label{sec:cp_method}
%Conformal prediction is a general framework, not a complete solution yet. To obtain the optimal prediction sets in context of KGE, we need to carefully design the conformal predictors tailored for KGE models. 
%to ensure the coverage property of the prediction sets (\textit{coverage desideratum}) for any choice of a nonconformity measure $S$ and any KGE model. Nonetheless, it does not provide a complete solution, as the quality of the resulting prediction sets can vary significantly with respect to \textit{size} and \textit{adaptiveness desideratum} depending on the design of $S$ and the way to construct prediction sets. 
%The most important ingredients for conformal predictors are nonconformity measure and the way to construct prediction sets. In this section, we propose several KGE-specific nonconformity measures and describe an efficient and easy-to-implement way to construct the prediction sets. 
To improve the basic set predictors, we apply conformal prediction, a general framework that requires adaptation to be effective in the context of KGE. The two essential components in this design are the nonconformity measure and the method for constructing answer sets. In this section, we propose several KGE-specific nonconformity measures and outline an efficient approach to constructing answer sets.
%To improve the basic set predictors, we apply conformal prediction. Conformal prediction provides a general framework but is not a standalone solution. To achieve optimal prediction sets in the context of KGE, conformal predictors must be specifically designed to align with the characteristics of KGE models. The two key components in this design are the nonconformity measure and the way to construct prediction sets. In this section, we propose several KGE-specific nonconformity measures and outline an efficient approach for constructing prediction sets.

\subsection{Nonconformity Measures}\label{sec:nonconform}
The probabilistic guarantees in Theorem \ref{th:coverage} hold under i.i.d assumption, regardless of the data distribution or the definition of the nonconformity measure. However, the size of the resulting answer sets depends on how effectively the nonconformity measure captures the underlying structure of the data. Next, we introduce several nonconformity measures for KGE models and explain the rationale behind each one.

Formally, given a set of training triples $\mathcal{T}_{1:n}$ and a test triple $t_{n+1}:= tr(q_{n+1}, e_{n+1})$, the nonconformity measure $S(\mathcal{T}_{1:n}, t_{n+1})$ estimates how unusual the triple $t_{n+1}$ is as a part of $\mathcal{T}_{1:n}$. 

\textbf{NegScore}. 
The underlying idea of KGE methods is to assign higher plausibility scores to positive triples and lower scores to negative triples. Therefore, a natural choice for the nonconformity score is the negative value of the plausibility score. The intuition here is that a lower plausibility score indicates a higher nonconformity, suggesting that the triple is less consistent with the existing triples represented in the training set. Formally, let $M_{\mathcal{T}_{1:n}}$ denote a KGE model trained on $\mathcal{T}_{1:n}$, then the corresponding nonconformity measure is defined as
%\todo{perhaps $\frac{1}{p}$ instead of $-p$ to have a non-negative measure? $\frac{1}{p} - 1 = \frac{1-p}{p}$ has an even nicer range}
\begin{equation}
    S(\mathcal{T}_{1:n}, t_{n+1}) = -M_{\mathcal{T}_{1:n}}(t_{n+1})
\end{equation}

\textbf{Minmax}. 
While the NegScore predictor directly uses the raw plausibility scores, the scale of these scores can vary significantly across different queries, potentially affecting the consistency and reliability of the nonconformity measure. To address this, we normalize the scores for each query using min-max scaling. This ensures that the nonconformity score reflects not only the raw plausibility but also the relative position of the triple within the score distribution for all possible triples in a given query. We then define the nonconformity measure as 
\begin{equation}
    S(\mathcal{T}_{1:n}, t_{n+1}) = -\overline M_{\mathcal{T}_{1:n}}(t_{n+1}),
\end{equation}
where
\begin{align}
    &\overline M(tr(q,e))=\\
    &\frac{M(tr(q,e))-\min_{e'\in E}M(tr(q,e'))}{\max_{e'\in E}M(tr(q,e'))-\min_{e'\in E}M(tr(q,e'))}.
\end{align}

\textbf{Softmax}. 
Another approach to normalizing plausibility scores is by using the softmax function, which converts the plausibility scores into a (uncalibrated) "probability distribution" over all possible answers for a given query. Unlike min-max scaling, Softmax scaling is more sensitive to the relative differences between scores, naturally highlighting the most likely triples while acknowledging others. This can result in more nuanced nonconformity measures, especially when the score distribution has varying degrees of separation between true and false triples. The nonconformity score is then defined as the of softmax outputs and the "ground truth" probability, which is assumed to be $1$ for the true answer. 
%Another idea is to convert the predictive scores into "probabilities" of each possible entities for each query, and define the distance of the estimated probability and the ground truth probability (i.e. $1$) as nonconformity score. 
\begin{equation}
    S(\mathcal{T}_{1:n}, t_{n+1}) = 1-\hat M_{\mathcal{T}_{1:n}}(t_{n+1}),
\end{equation}
where
\begin{equation}
    \hat M(tr(q,e))=\frac{\exp(M(tr(q,e)))}{\sum_{e'\in E}\exp{(M(q,e'))}}.
\end{equation}

\subsection{Answer Set Construction} \label{sec:construction}
% In order to have the probabilistic guarantee in theorem \ref{th:coverage}, we construct the prediction sets according to equation \ref{eq:construct}, namely, we include all entity $e\in E$ such that
% \begin{equation}\label{eq:construct_kg}
%     \frac{|\{i=1,\dots,n+1:\alpha_i\geq\alpha_{n+1}\}|}{n+1}>\epsilon,
% \end{equation}
% where
% \begin{align*}
%     &\alpha_i:=S(\mathcal{T}_{1:n}\cup \{tr(q_{n+1}, e)\}, tr(q_i, e_i)), \\
%     &i=1,\dots,n\\
%     &\alpha_{n+1}:=S(\mathcal{T}_{1:n}\cup \{tr(q_{n+1}, e)\}, tr(q_{n+1}, e)),
% \end{align*}
% \todo[inline]{shorten above}
If we construct answer sets as describe in Section \ref{sec:pre}, we need to retrain the KGE model with $\mathcal{T}_{1:n}\cup \{tr(q_{n+1}, e)\}$ and recalculate the nonconformity scores of training triples for testing each triple $tr(q_{n+1}, e)$ (for all $e\in E$). It is computationally not feasible for KGE methods, given the huge number of possible entities $e\in E$ and the time-consuming training and hyper-parameter tuning process. 

We adopt so called split/inductive conformal prediction \cite{vovk2005algorithmic, lei2015conformal} to address this issue (see Algorithm \ref{algo:cp_base} for details). The training set of size $n$ is first divided into a proper training set $\mathcal{T}_{1:m}$ of size $m<n$ and a calibration set $\mathcal{T}_{m+1:n}$ of size $n-m$. Rather than using the entire training set to train the KGE model and evaluate nonconformity scores, we train the KGE model once on the proper training set $\mathcal{T}_{1:m}$ and use it to calculate the nonconformity scores on the calibration set $\mathcal{T}_{m+1:n}$. Intuitively, if the calibration set is chosen randomly and is sufficiently large, its empirical coverage should closely match the true coverage probability for a new query. This strategy significantly increases the efficiency of the conformal predictors while preserving the probabilistic guarantees in Theorem \ref{th:coverage} \cite{lei2018distribution}. 

Formally, in split conformal prediction, if we construct answer sets by including all entity $e\in E$ such that

\begin{equation}\label{eq:construct_split}
    \frac{|\{i=m+1,\dots,n+1:\alpha_i\geq\alpha_{n+1}\}|+1}{n-m+1}>\epsilon,
\end{equation}
where
\begin{align*}
    &\alpha_i:=S(\mathcal{T}_{1:m}, tr(q_i, e_i)), i=m+1,\dots,n\\
    &\alpha_{n+1}:=S(\mathcal{T}_{1:m}, tr(q_{n+1}, e)).
\end{align*}
%It is proven in \cite[Lemma 2.1]{lei2015conformal} and \cite[chapter 4.2.2]{vovk2005algorithmic} that the coverage guarantee in Theorem \ref{th:coverage} still holds if we can identify a nonconformity measure that is invariant with respect to the permutation of its first argument (the training set). 
%The following proposition can be derived based on Theorem \ref{th:coverage} and \cite[Appendix A]{lei2018distribution}.
Based on Theorem 2.2 in \citet{lei2018distribution}, we have the following corollary:
\begin{corollary}\label{prop:cover}
    Given a set of triples $\mathcal{T}_{1:n+1}$ that are i.i.d, an error rate $\epsilon\in (0,1)$ and any nonconformity measure $S$. If $n$ is large, the answer set of a test query $\hat C(q_{n+1})$ constructed following Equation \ref{eq:construct_split} satisfies
    \begin{equation}\label{eq:prop_lower}
        \mathbb{P}(e_{n+1}\in\hat C(q_{n+1}))\geq 1-\epsilon
    \end{equation}
    furthermore, if there are no ties between nonconformity scores in the calibration set $\mathcal{T}_{m+1:n}$, we have
    \begin{equation}\label{eq:prop_upper}
        \mathbb{P}(e_{n+1}\in\hat C(q_{n+1}))\leq 1-\epsilon + \frac{1}{n-m+1}
    \end{equation}
\end{corollary}
%The detailed proof can be found in Appendix \ref{app:proof}. 
%Intuitively, by the i.i.d assumption, the rank of the nonconformity scores $\alpha_{m+1}, \dots\alpha_{n+1}$ is uniform among $1,2,\dots, n-m+1$. Therefore, with probability at least $1-\epsilon$, $e_{n+1}$ falls in $\hat C(q_{n+1})$. 
The proof of Proposition \ref{prop:cover} can be found in \cite[Appendix A.1]{lei2018distribution} \footnote{Unlike \cite{lei2018distribution}, we split the training set unevenly, resulting in slight differences in Equation \ref{eq:prop_upper}.}.
Note that the additional assumption for Equation \ref{eq:prop_upper} is a quite weak assumption, by using a random tie-breaking rule, this assumption could be avoided entirely. %As confirmed by the experiments in Table \ref{tab:main1_filter}, \ref{tab:main2_filter}, \ref{tab:main1} and \ref{tab:main2}, the coverage of conformal predictors is highly concentrated around $1-\epsilon$.

\begin{algorithm}[h!]
\caption{Pseudocode for Split Conformal Prediction.}\label{algo:cp_base}
\begin{algorithmic}[1]
    \Require A training set $\mathcal{T}_{1:m}$, a calibration set $\mathcal{T}_{m+1:n}$, a testing query $q_{n+1}$, an error rate $\epsilon$ and a nonconformity measure $S$.
    \State
    \State $\triangleright$ Calibration Step
    \State $L'\gets$ an empty set
    \For{\textbf{each triple} $tr(q',e')$ \textbf{in} $\mathcal{T}_{m+1:n}$} 
        \State $L'\gets L'\cup \{S(\mathcal{T}_{1:m}, tr(q',e'))\}$
    \EndFor
    \State $t\gets\lceil(|\mathcal{T}_{m+1:n}|+1)(1-\epsilon)/|\mathcal{T}_{m+1:n}|\rceil$
    \State $\tau\gets t$ th quantile of elements in $L'$
    \State
    \State $\triangleright$ Prediction Step
    \State $L\gets$ an empty set
    \For{\textbf{each entity} $e$ \textbf{in} $E$}
        \State $L\gets L\cup \{(e, S(\mathcal{T}_{1:m}, tr(q_{n+1},e))\}$
    \EndFor

    \State $\hat C(q_{n+1})\gets$ an empty set
    \For{$(e, s)$ \textbf{in} $L$}
        \If{$s<\tau$}
            \State $\hat C(q_{n+1})\gets \hat C(q_{n+1})\cup \{e\}$
        \EndIf
    \EndFor
    \State \textbf{return} $\hat C(q_{n+1})$
\end{algorithmic}
\end{algorithm}

\subsection{Time Complexity Analysis}
The use of split conformal prediction in Section \ref{sec:construction} significantly reduces the computational effort by leveraging a pre-trained KGE model, eliminating the need to train a KGE model for each $tr(q_{test}, e)$ to recompute nonconformity scores.
This approach ensures that the scalability of our method is independent of the training graph size. In this section, we analyze the time complexity of the proposed method.

Let $T$ represent the time required to compute nonconformity scores. 
For a calibration set $\mathcal{T}{m+1:n}$ of size $n-m$ and a test set $\mathcal{T}{1:j}$ of size $j$, the time complexity of split conformal prediction is:
\begin{align}
    \mathcal{O}\bigg((n-m+j)T+(n&-m)\log(n-m)\\
    &+j\log(n-m)\bigg)\notag.
\end{align}
Here, the term $(n-m)\log(n-m)$ accounts for sorting the nonconformity scores in the calibration set, while $\log(n-m)$ for finding the rank of a test nonconformity score. Since $T$ scales linearly with the number of entities $|E|$, the complexity can be expressed as:
\begin{align}
    \mathcal{O}\bigg((n-m+j)|E|+(n&-m)\log(n-m)\\
    &+j\log(n-m)\bigg)\notag.
\end{align}
As $n-m$ and $j$ are independent of $|E|$ and typically small, the asymptotic complexity for large $|E|$ becomes $\mathcal{O}(|E|)$. This result highlights the efficiency and scalability of the method for larger KGs.

\begin{table*}[h!]
\resizebox{\textwidth}{!}{%
\begin{tabular}{ccccc|ccccc}
\toprule[1.5pt]
\multicolumn{5}{c}{WN18} & \multicolumn{5}{c}{FB15k} \\
model & MR & methods & coverage & size & model & MR & methods & coverage & size \\\midrule[1.2pt]

 &  & naive & {\color[HTML]{FF0000} 0.44 (0.004)} & 12.28 (1.262) &  &  & naive & {\color[HTML]{FF0000} 0.73 (0.001)} & 258.14 (0.834) \\
 &  & Platt & {\color[HTML]{FF0000} 0.85 (0.002)} & 4043.41 (89.765) &  &  & Platt & {\color[HTML]{FF0000} 0.84 (0.002)} & 1197.64 (45.923) \\
 &  & topk & 0.90 (0.000) & 48.01 (0.739) &  &  & topk & 0.90 (0.000) & 336.68 (1.332) \\
 &  & {\ul NegScore} & 0.90 (0.001) & \textbf{20.99 (0.587)} &  &  & {\ul NegScore} & 0.90 (0.001) & \textbf{45.18 (0.280)} \\
 &  & {\ul Softmax} & 0.90 (0.001) & 112.80 (4.650) &  &  & {\ul Softmax} & 0.90 (0.000) & 414.88 (2.390) \\
\multirow{-6}{*}{TransE} & \multirow{-6}{*}{245.82 (6.368)} & {\ul Minmax} & 0.90 (0.001) & 113.57 (5.098) & \multirow{-6}{*}{TransE} & \multirow{-6}{*}{43.05 (0.176)} & {\ul Minmax} & 0.90 (0.000) & 275.27 (3.217) \\\midrule

 &  & naive & 0.91 (0.003) & 17690.17 (117.856) &  &  & naive & {\color[HTML]{FF0000} 0.71 (0.003)} & 748.09 (5.216) \\
 &  & Platt & 0.90 (0.002) & 16950.17 (116.477) &  &  & Platt & {\color[HTML]{FF0000} 0.88 (0.002)} & 1156.33 (7.552) \\
 &  & topk & 0.90 (0.001) & 50.85 (2.440) &  &  & topk & 0.90 (0.000) & 408.43 (3.752) \\
 &  & {\ul NegScore} & 0.90 (0.002) & \textbf{1.27 (0.010)} &  &  & {\ul NegScore} & 0.90 (0.001) & \textbf{52.31 (0.605)} \\
 &  & {\ul Softmax} & 0.90 (0.001) & 1.91 (0.249) &  &  & {\ul Softmax} & 0.90 (0.000) & 140.36 (3.196) \\
\multirow{-6}{*}{RotatE} & \multirow{-6}{*}{478.13 (44.173)} & {\ul Minmax} & 0.90 (0.003) & 3.88 (0.698) & \multirow{-6}{*}{RotatE} & \multirow{-6}{*}{61.77 (0.976)} & {\ul Minmax} & 0.90 (0.001) & 42.35 (1.064) \\\midrule

 &  & naive & {\color[HTML]{FF0000} 0.58 (0.011)} & 300.09 (24.737) &  &  & naive & {\color[HTML]{FF0000} 0.58 (0.016)} & 121.14 (13.125) \\
 &  & Platt & {\color[HTML]{FF0000} 0.80 (0.005)} & 2021.25 (223.393) &  &  & Platt & {\color[HTML]{FF0000} 0.87 (0.003)} & 615.92 (24.759) \\
 &  & topk & 0.90 (0.001) & 54.46 (1.640) &  &  & topk & 0.90 (0.000) & 394.18 (2.357) \\
 &  & {\ul NegScore} & 0.91 (0.001) & 45.50 (3.630) &  &  & {\ul NegScore} & 0.90 (0.001) & 168.64 (4.506) \\
 &  & {\ul Softmax} & 0.90 (0.001) & 2.14 (0.062) &  &  & {\ul Softmax} & 0.90 (0.000) & \textbf{72.61 (0.491)} \\
\multirow{-6}{*}{RESCAL} & \multirow{-6}{*}{321.73 (21.501)} & {\ul Minmax} & 0.90 (0.002) & \textbf{2.02 (0.075)} & \multirow{-6}{*}{RESCAL} & \multirow{-6}{*}{65.52 (1.815)} & {\ul Minmax} & 0.90 (0.001) & 79.62 (5.573) \\\midrule

 &  & naive & {\color[HTML]{FF0000} 0.47 (0.002)} & 36.54 (7.024) &  &  & naive & {\color[HTML]{FF0000} 0.35 (0.013)} & 19.47 (0.897) \\
 &  & Platt & {\color[HTML]{FF0000} 0.84 (0.001)} & 1265.82 (205.375) &  &  & Platt & 0.90 (0.001) & 485.01 (2.341) \\
 &  & topk & 0.90 (0.001) & 57.48 (2.439) &  &  & topk & 0.90 (0.000) & 362.91 (1.414) \\
 &  & {\ul NegScore} & 0.90 (0.001) & 2244.87 (405.033) &  &  & {\ul NegScore} & 0.90 (0.000) & 156.02 (2.872) \\
 &  & {\ul Softmax} & 0.90 (0.001) & 2.02 (0.047) &  &  & {\ul Softmax} & 0.90 (0.000) & \textbf{28.40 (0.421)} \\
\multirow{-6}{*}{DistMult} & \multirow{-6}{*}{370.21 (20.313)} & {\ul Minmax} & 0.90 (0.002) & \textbf{1.51 (0.049)} & \multirow{-6}{*}{DistMult} & \multirow{-6}{*}{45.13 (0.556)} & {\ul Minmax} & 0.90 (0.000) & 40.35 (0.894) \\\midrule

 &  & naive & 0.94 (0.007) & 19968.55 (142.153) &  &  & naive & {\color[HTML]{FF0000} 0.28 (0.011)} & 50.24 (1.309) \\
 &  & Platt & {\color[HTML]{FF0000} 0.86 (0.002)} & 16788.25 (125.356) &  &  & Platt & 0.90 (0.001) & 922.43 (9.375) \\
 &  & topk & 0.91 (0.002) & 40.58 (1.811) &  &  & topk & 0.90 (0.000) & 414.06 (4.429) \\
 &  & {\ul NegScore} & 0.90 (0.003) & 1.47 (0.008) &  &  & {\ul NegScore} & 0.90 (0.001) & 99.11 (9.237) \\
 &  & {\ul Softmax} & 0.90 (0.001) & \textbf{1.10 (0.007)} &  &  & {\ul Softmax} & 0.90 (0.001) & \textbf{37.20 (0.927)} \\
\multirow{-6}{*}{ComplEx} & \multirow{-6}{*}{454.21 (27.914)} & {\ul Minmax} & 0.90 (0.002) & 2.83 (0.171) & \multirow{-6}{*}{ComplEx} & \multirow{-6}{*}{66.87 (1.603)} & {\ul Minmax} & 0.90 (0.001) & 118.43 (20.418) \\\midrule

 &  & naive & {\color[HTML]{FF0000} 0.50 (0.015)} & 3.42 (1.423) &  &  & naive & {\color[HTML]{FF0000} 0.45 (0.026)} & 819.05 (111.870) \\
 &  & Platt & {\color[HTML]{FF0000} 0.84 (0.007)} & 79.09 (1.785) &  &  & Platt & {\color[HTML]{FF0000} 0.88 (0.014)} & 4824.48 (699.253) \\
 &  & topk & 0.90 (0.001) & 53.21 (1.475) &  &  & topk & 0.90 (0.000) & 386.02 (3.107) \\
 &  & {\ul NegScore} & 0.90 (0.002) & 1.50 (0.067) &  &  & {\ul NegScore} & 0.90 (0.000) & \textbf{44.39 (2.225)} \\
 &  & {\ul Softmax} & 0.90 (0.001) & 1.57 (0.090) &  &  & {\ul Softmax} & 0.90 (0.000) & 79.30 (6.029) \\
\multirow{-6}{*}{ConvE} & \multirow{-6}{*}{311.27 (12.598)} & {\ul Minmax} & 0.90 (0.001) & \textbf{1.34 (0.028)} & \multirow{-6}{*}{ConvE} & \multirow{-6}{*}{67.56 (1.644)} & {\ul Minmax} & 0.90 (0.000) & 48.31 (8.576)\\\bottomrule[1.5pt]
\end{tabular}%
}
\caption{Quality of the filtered answer sets on WN18 and FB15k datasets. This table presents the mean rank (MR) of KGE models (lower is better), along with the coverage and size of answer sets generated using various set predictors. Conformal predictors are underlined. Means and standard deviations (in the brackets) over 15 trials are reported at the $10\%$ level ($\epsilon=0.1$). Predictors that fail to meet the coverage threshold of $1-\epsilon$ (0.9) are highlighted in red. The best predictors, which meet the coverage desideratum and minimize answer set size, are highlighted in bold.}
\label{tab:main1_filter}
\end{table*}

\section{Experiments}\label{sec:exp}
In this section, we present five experiments that evaluate the quality of the answer sets from (baseline) predictors in Section \ref{sec:baselines} (\textit{naive}, \textit{Platt}, \textit{topk}) and conformal predictors (\textit{NegScore}, \textit{Softmax}, \textit{Minmax}) in Section \ref{sec:cp_method} based on coverage, size and adaptiveness desiderata. 

\textbf{Datasets and Backbone KGE Methods.} In our experiments, we use four commonly used benchmark link prediction datasets: WN18 \cite{bordes2013translating}, WN18RR \cite{dettmers2018convolutional}, FB15k \cite{bordes2013translating} and FB15k237 \cite{toutanova2015observed} and six representative KGE methods: TransE \cite{bordes2013translating}, RotatE \cite{sun2019rotate}, RESCAL \cite{nickel2011rescal}, DistMult \cite{yang2015distmult}, ComplEx \cite{trouillon2016complex} and ConvE \cite{dettmers2018convolutional}. We provide more information about the experimental settings in Appendix \ref{app:kge_models}. 

\textbf{Evaluation Metrics.} 
A good set predictor should cover the true answer with a probability of at least $1-\epsilon$ (coverage desideratum) and provide smaller sets (size desideratum). 
Let $\mathcal{T}_{1:j}$ be a set of triples disjoint with $\mathcal{T}_{1:n}$ as test set. We measure the following \emph{empirical coverage probability} to verify the coverage desideratum, i.e., $\mathbb{P}(e_{n+1}\in\hat{C}(q_{n+1}))\geq1-\epsilon$:
\begin{equation}
    \frac{1}{|\mathcal{T}_{1:j}|}\sum_{tr(q,e)\in\mathcal{T}_{1:j}}\mathbbm{1}[e\in\hat C(q)]
\end{equation}
We measure the \emph{average size} of the answer sets as following:
\begin{equation}
    \frac{1}{|\mathcal{T}_{1:j}|}\sum_{tr(q,e)\in\mathcal{T}_{1:j}}|\hat C(q)|
\end{equation}

\subsection{Experiment 1: Coverage and Set Size on WN18 and FB15k}
%\todo[inline]{add calibration baselines}
%\todo{perhaps add something like top10, top100 to the baselines?}
% A good set predictor should cover the true answer with a probability of at least $1-\epsilon$ (coverage desideratum) and predict smaller sets (size desideratum). 
In this experiment, we evaluate desiderata by measuring the empirical coverage probability ($coverage$) and the average size ($size$) of answer sets for each method. 
Each procedure is repeated 15 times, and we report the mean and standard deviation (in brackets) across trials in Table \ref{tab:main1_filter}.
As usual in the evaluation of link prediction, for each query, 
we consider only answer candidates that did not already occur in the training and validation data.

As demonstrated in Proposition \ref{prop:cover}, conformal predictors consistently meet the coverage desideratum, with coverage tightly concentrated around $1-\epsilon$.
Compared to baseline predictors that also satisfy the coverage desideratum, conformal predictors outperform them in terms of producing smaller answer sets, thus better satisfying the size desideratum. 

The naive predictor, on the other hand, frequently fails to meet the coverage desideratum, often providing lower coverage than necessary, indicating that the plausibility scores are generally overconfident. While applying a calibration technique to the naive predictor (Platt predictor) improves coverage, it still does not meet the coverage guarantee, and the resulting significant increase in set size makes it impractical for use. 

The topk predictor meets the coverage desideratum but generally produces larger and fixed-sized answer sets compared to the conformal predictors.
In Appendix \ref{app:fixsize}, we also discuss simpler fixed-sized predictors and compare them to the topk predictor.
It is worth noting that the topk predictor can be viewed as a specific case of the conformal predictor, where the nonconformity score is defined by the rank position.
%\todo[inline]{describe the results of fixed-size predictors}

Additionally, we observed that there is no universally optimal nonconformity score for conformal predictors; the choice is model- and dataset-dependent. For instance, NegScore seems to better capture the nonconformity of triples in distance-based models (TransE, RotatE), while Softmax and Minmax scores are more suitable for semantic matching models (RESCAL, DistMult, and ComplEx). 

The calibration technique in \cite{Tabacof2020calibration, Safavi2020calibration} should theoretically enhance the design of the nonconformity measure and thereby improve the conformal predictor. However, in our setting, it fails to do so. The results are presented in Table \ref{tab:cali_main}, with a discussion of potential reasons provided in the Appendix \ref{app:cali_cp}.

Due to space constraints, additional results, including those without filtering existing answers and results from more datasets, are provided in Appendix \ref{app:cover}. The conclusions are consistent across all scenarios.

\subsection{Experiment 2: Adaptiveness of Answer Sets}
\begin{figure*}[h!]
    \centering
    \includegraphics[width=\linewidth]{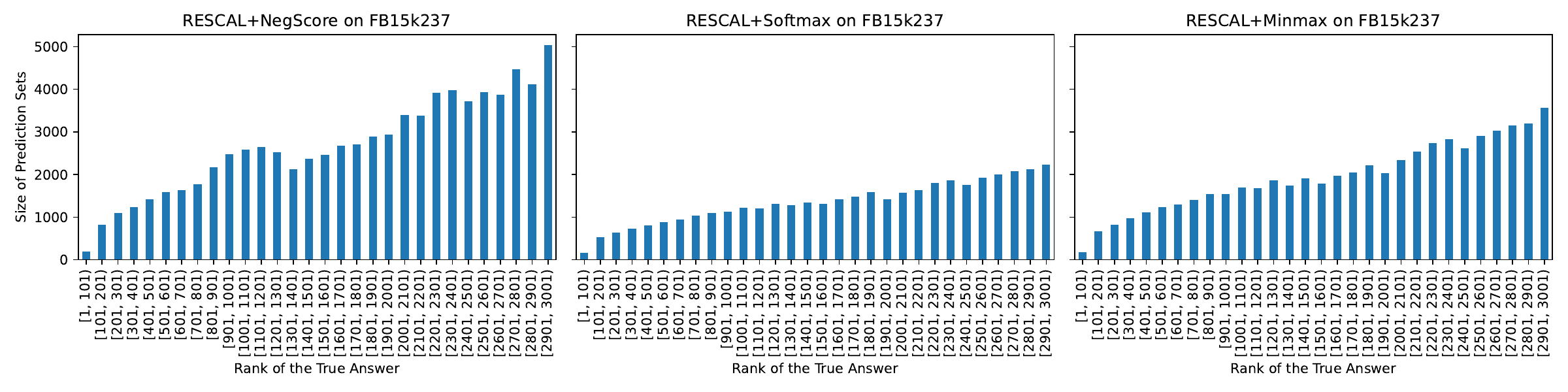}
    \caption{This figure shows the size of answer sets stratified by the difficulty level of queries. It shows the adaptiveness of different conformal predictors (built on RESCAL models) on the FB15k237 dataset, more results can be found in Figure \ref{fig:adap_TransE_FB15k} - \ref{fig:adap_ConvE_FB15k237} in Appendix. }
    \label{fig:adap_RESCAL_fb15k237}
\end{figure*}

This experiment aims to determine whether the size of answer sets adapts well to the
difficulty of the query. 
Unfortunately, there is no well justified way
to evaluate query difficulty at the moment.
We therefore follow the experimental protocol
used in \cite{Angelopoulos2021image}
for computer vision tasks. The authors evaluate 
difficulty by looking at the rank of the true
label in the ranking obtained from the classifier by ordering labels according to
their softmax-probabilities. The higher the
rank, the more difficult the query.
Analogously, we use the rank of the true answer
given by the KGE model 
to evaluate query difficulty. 

% A lower-ranked true answer may result from either aleatoric uncertainty, due to multiple correct answers, or epistemic uncertainty, reflecting the model's limitations in learning and generalization. 

We categorize queries by difficulty levels based on the rank of the true answer (e.g., 1-100, 101-200, etc.). For each difficulty level, we calculate the average size of answer sets. Figure \ref{fig:adap_RESCAL_fb15k237} illustrates the size of answer sets stratified by query difficulty. The x-axis represents rank intervals from 1 to 3000, segmented into 100-rank bins (reflecting different difficulty levels), while the y-axis shows the average size of answer sets within each interval.

We observe that the size of answer sets generated by conformal predictors closely aligns with the difficulty levels of the queries, thereby fulfilling the adaptiveness desideratum. This is a valuable property because, in practice, the true answer to a query is unknown. By examining the size of the answer set, we can estimate the predictive uncertainty for the query.

\subsection{Experiment 3: Impact of Calibration Set Size on Answer Set Quality}
In this experiment, we investigate the impact of size of the calibration set, $\mathcal{T}_{m+1:n}$, on the quality of answer sets in terms of coverage and size desiderata. 
We randomly sampled calibration sets of 10, 100, 200, and 500 triples from the validation set for use in conformal prediction. We then evaluated the coverage and average size of the resulting answer sets. This process was repeated 20 times to compute the mean and standard deviation of the results. For comparison, we also evaluated the answer sets generated using the entire validation set as the calibration set.

As shown in Proposition \ref{prop:cover}, the coverage of conformal predictors with an i.i.d calibration set should fall between $1-\epsilon$ and $1-\epsilon+\frac{1}{n-m}$, where $n-m$ is the size of the calibration set. This is confirmed by the results in Table \ref{tab:efficiency}.
The size of answer sets generated by split conformal predictors depends on the alignment between the distribution of nonconformity scores in the calibration set and those in the original training set (which includes both the proper training set and the calibration set). A larger calibration set typically better represents the original training set, leading to tighter answer sets, as confirmed by the results in Table \ref{tab:efficiency}.
Notably, even with a relatively small calibration set, the quality of the answer sets closely approximates that obtained using the entire validation set. 

%These findings suggest that conformal predictors can produce high-quality prediction sets with a small number of calibration triples. %\todo{perhaps your point here is not that a sufficiently large set is required to obtain the guarantees, but that even small sets can give good guarantees? Perhaps rewrite a little bit? I guess, it's somewhat related to evaluating "sample efficiency"}.

\begin{table}[h!]
\resizebox{.48\textwidth}{!}{%
\begin{tabular}{ccll}
\toprule
Predictor & Size of Calibration Set & Coverage & Size \\ \midrule
\multirow{5}{*}{NegScore} & 10   & 0.98 (0.018)       & 7626.93 (6145.132)   \\
& 100   & 0.91 (0.019)       & 54.67 (18.268)   \\
& 200   & 0.91 (0.014)       & 50.88 (12.725)    \\
& 500   & 0.90 (0.010)    & 47.58 (6.891)   \\
& Entire validation set   & 0.90 (-)      & 46.26 (-)   \\ \midrule
\multirow{5}{*}{Softmax} & 10   &  0.97 (0.002)      & 8604.95 (7395.591)   \\
& 100   & 0.91 (0.018)       & 2.70 (1.158)   \\
& 200   & 0.91 (0.012)       & 2.31 (0.591)    \\
& 500   & 0.90 (0.008)    & 2.07 (0.255)   \\
& Entire validation set    & 0.90 (-)      &  2.07 (-)  \\ \midrule
\multirow{5}{*}{Minmax} & 10   &   0.96 (0.03)     & 2796.73 (546.911)   \\
& 100   & 0.91 (0.020)       & 2.24 (0.671)   \\
& 200   & 0.91 (0.016)       & 2.18 (0.390)    \\
& 500   & 0.90 (0.009)    & 1.99 (0.145)   \\
& Entire validation set    & 0.90 (-)      &  1.94 (-)  \\\bottomrule
\end{tabular}%
}
\caption{This table shows the coverage and size (with means and standard deviations over 20 trials) of answer sets generated by different predictors using varying sizes of calibration sets on the WN18 dataset. }
\label{tab:efficiency}
\end{table}

\subsection{Experiment 4: Impact of Error Rate on Answer Set Quality}
\begin{figure}[h!]
    \centering
    \includegraphics[width=0.87\linewidth]{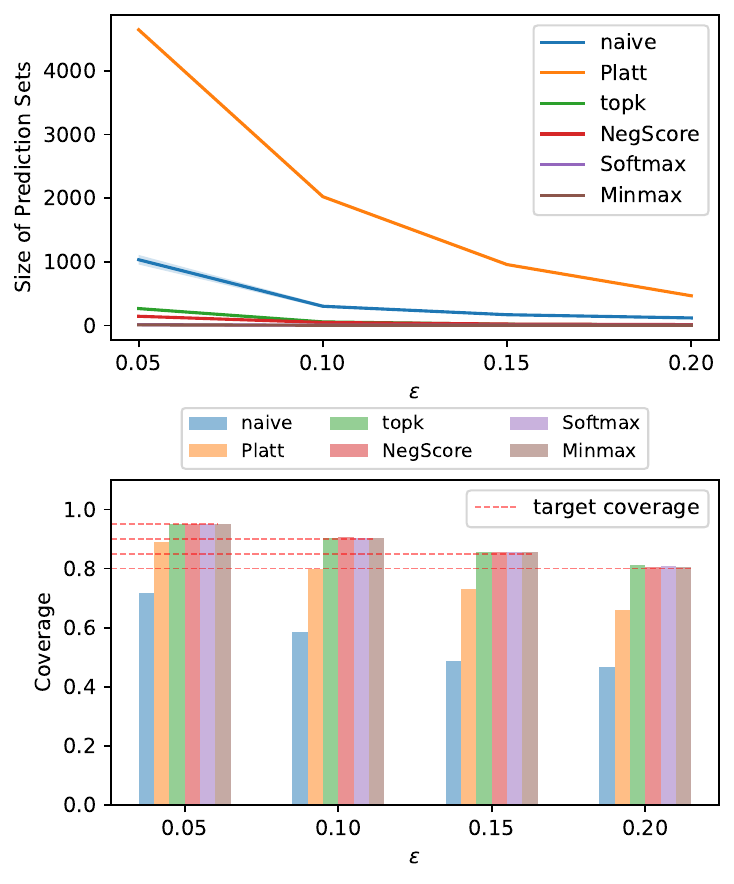}
    \caption{This figure shows how the coverage and size of answer sets change with respect to $\epsilon$ across different predictors on the WN18 dataset.}
    \label{fig:error_rate}
\end{figure}

In this experiment, we examine the effect of the user-specified error rate ($\epsilon$) on the quality of answer sets. Figure \ref{fig:error_rate} illustrates how $\epsilon$ influences the size of answer sets (upper diagram) and the coverage of answer sets (lower diagram) across various predictors. The red line in lower diagram correspond to the desired coverage $1-\epsilon$. 

As expected, the size of answer sets decreases as $\epsilon$ increases, aligning with the requirements discussed in Section \ref{sec:pre}. The topk and conformal predictors consistently generate smaller answer sets compared to the naive and Platt predictors. Notably, conformal predictors produce the most compact answer sets when the error rate is set to a very low value. In terms of coverage, conformal predictors consistently meet the probabilistic guarantee in Proposition \ref{prop:cover} across the range of $\epsilon$ values.
%\todo{perhaps write explicitly that the red lines correspond to $1-\epsilon$ in Figure \ref{fig:error_rate}?} 

\section{Discussion}
%\todo[inline]{discuss the logic of quantify uncertainty in our paper: first satisfy guarantee and then look for suitable nonconformity measure}
Our method predicts answer sets for queries with a guaranteed coverage of the true answer at a pre-specified probability, such as $90\%$, while maintaining a small average size. Unlike ranking-based outputs, our approach is particularly well-suited for decision-making in high-stakes domains, including medical diagnosis, drug discovery, and fraud detection. For instance, a doctor could use our method to automatically eliminate a large number of irrelevant diseases, thereby referring the patient to the most appropriate specialists. Additionally, our method is easy to implement and is compatible not only with any KGE models but also with embedding methods capable of answering more complex queries \cite{ren2020query2box, he2024generating, he2024dage} and approximate statistical reasoning in ontology \cite{zhu2023towards, zhu2024approx}.

The adaptability of our answer sets to the uncertainty of queries also enables our method to quantify the predictive uncertainty of KGE models. This feature broadens the applicability of our approach by systematically identifying hard or uncertain queries during testing. Detecting such queries can help identify potential failure cases or outliers, alerting users when the model's predictions may be unreliable.

\section{Limitations}
A limitation of our method is the requirement to divide the training set into two parts: one for training the model and another for calculating the nonconformity scores, due to the adoption of split conformal prediction. This division reduces the number of triples available for model training. However, this issue is mitigated by the fact that the validation set, typically reserved for hyperparameter tuning, can also serve as the calibration set. Moreover, as demonstrated in Experiment 4, even a small subset of the validation set is sufficient to produce nearly optimal answer sets.

Another limitation is that the probabilistic guarantee provided by Theorem \ref{th:coverage} and Proposition \ref{prop:cover} relies on the i.i.d. assumption, which may not hold under distribution shifts. We are currently extending our conformal predictors to covariant shift case, where only the input distribution $P(X)$ changes while the conditional distribution $P(Y|X)$ remains the same. We begin with the simpler scenario where the likelihood ratio between the training and test distributions is known. Following \cite{tibshirani2019conformal}, we weight each nonconformity score proportionally to the likelihood ratio to ensure the probabilistic guarantee in Proposition \ref{prop:cover} holds beyond the i.i.d. assumption.

%\todo{really very nice work :)}

\section{Acknowledgements}
The authors thank the International Max Planck Research School for Intelligent Systems (IMPRS-IS) for supporting Yuqicheng Zhu, Bo Xiong and Yunjie He. The work was partially supported by the Horizon Europe projects EnrichMyData (Grant Agreement No.101070284), Graph-Massivier (Grant Agreement No.101093202) and Dome 4.0 (Grant Agreement No.953163).

\newpage
\bibliography{anthology}

\newpage
\appendix

\section{Baseline Predictors}\label{app:code}
\subsection{Naive Predictor}
We provide detailed pseudocode for naive predictor in this section. See Algorithm \ref{algo:naive} for details.
\begin{algorithm}[h!]
\caption{Pseudocode for naive predictor.}\label{algo:naive}
\begin{algorithmic}[1]
    \Require A KGE model $M_\theta$ trained on $\mathcal{T}_{1:n}$, a testing query $q_{n+1}$ and an error rate $\epsilon$.
    \State $L\gets$ an empty set
    \For{\textbf{each entity} $e$ \textbf{in} $E$}
        \State $L\gets L\cup \{(e, M_\theta(tr(q_{n+1},e)))\}$
    \EndFor
    \State normalize the scores in $L$ with softmax function.
    \State $\overline L\gets$ sort elements in $L$ based on normalized scores (from largest to smallest).
    \State $p\gets 0$
    \State $\hat C(q_{n+1})\gets$ an empty set
    \For{$(e, s)$ \textbf{in} $\overline L$}
        \State $p\gets p+s$
        \If{$p<1-\epsilon$}
            \State $\hat C(q_{n+1})\gets \hat C(q_{n+1})\cup \{e\}$
        \EndIf
    \EndFor
    \State \textbf{return} $\hat C(q_{n+1})$
\end{algorithmic}
\end{algorithm}

\subsection{Platt Predictor}\label{app:platt}
The Platt predictor enhances the naive predictor using a calibration technique. The only difference in its procedure, as outlined in line 5 of Algorithm \ref{algo:naive}, is the modification of softmax outputs through temperature scaling\cite{guo2017calibration} — a multiclass extension of Platt scaling \cite{platt1999probabilistic}.

Temperature scaling employs a single scalar parameter $T > 0$ across all possible answer entities for a given query. Let $M_\theta(tr(q_{n+1}), e_i)$ represent the plausibility score of entity $e_i\in E$ for query $q_{n+1}$. The calibrated score $\hat s_i$ is then calculated as 
\begin{equation}\label{eq:calibration}
    \hat s_i = \sigma(M_\theta(tr(q_{n+1}), e_i)/T),
\end{equation}
where $\sigma(\cdot)$ is the softmax function. 

The parameter $T$, known as the temperature, "softens" the softmax output by increasing its entropy when $T > 1$. As $T \rightarrow \infty$, the probability $\hat{s}_i$ approaches $1/|E|$, indicating maximum uncertainty. When $T=1$, the original softmax output is recovered. Conversely, as $T \rightarrow 0$, the probability collapses to a point mass ($\hat{s}_i = 1 $). The optimal value of $T$ is determined by minimizing the negative log-likelihood on the validation set.
\begin{equation}
    NLL = -\frac{1}{N}\sum_{i=1}^{N}\log (\frac{\hat s_i}{\sum_{j=1}^{N}\hat s_j}),
\end{equation}
where $N$ is the size of calibration (validation) set.

\subsection{Fixed-sized Predictor}\label{app:fixsize}
The ranking-based metric $Hits@K$ evaluate how often KGE models place the correct answers within the top-K entities, implicitly suggesting that the top-K entities should be chosen as answer sets. Based on $Hits@K$, we evaluate the quality of \emph{fixed-sized} set predictor, which produce top-K entities (with a manually chosen $K$) as the answer set. 
%However, the number of correct answers varies by query. For example, a query asking for a country’s capital has only one correct answer, while a query about students supervised by a professor may have many. Top-K entities as answer sets can underrepresent queries with more than $K$ correct answers and overrepresent those with fewer than $K$.

We select $K$ values commonly used in Hits@K metrics (1, 3, 10, 100). The results in Table \ref{tab:fixsize_filter} and \ref{tab:fixsize} demonstrate that coverage is highly sensitive to the choice of $K$. Concretely, the fixed-sized set predictor either fails to meet the coverage desideratum or generate unnecessarily large answer sets. 

Consequently, in the main body of the paper, we adopt the topk predictor, where we use $K$ that cover the true answer in $1-\epsilon$ of queries in the validation set. The topk predictor effectively balances the trade-off between coverage and average size. However, unlike conformal predictors, the topk predictor cannot adapt answer set sizes to the difficulty of individual queries, as it uses the same size for all queries.

\begin{table}[t!]
\resizebox{\linewidth}{!}{%
\begin{tabular}{lcl|lcl}
\toprule
\multicolumn{3}{c}{WN18} & \multicolumn{3}{c}{FB15k} \\
method & coverage & size & method & coverage & size\\\midrule
top1 & 0.45 & 0.48 & top1 & 0.12 & 0.17\\
top3 & 0.65 & 1.76 & top3 & 0.24 & 0.71\\
top10 & 0.80 & 7.32 & top10 & 0.46 & 3.91\\
top100 & 0.92 & 90.91 & top100 & 0.79 & 71.46\\
topk & 0.90 & 52.96 & topk & 0.90 & 395.11\\
\bottomrule
\end{tabular}%
}
\caption{Comparison of the fixed-size set predictors and the top-K predictor with $K$ learned based on the validation set. The table reports the mean values of coverage and the average size of filtered answer sets over 15 trials. Results are based on the RESCAL model applied to the WN18 and FB15k datasets.}
\label{tab:fixsize_filter}
\end{table}

\begin{table}[t!]
\resizebox{\linewidth}{!}{%
\begin{tabular}{lcl|lcl}
\toprule
\multicolumn{3}{c}{WN18} & \multicolumn{3}{c}{FB15k} \\
method & coverage & size & method & coverage & size\\\midrule
top1 & 0.45 & 1.00 & top1 & 0.12 & 1.00\\
top3 & 0.65 & 3.00 & top3 & 0.24 & 3.00\\
top10 & 0.80 & 10.00 & top10 & 0.46 & 10.00\\
top100 & 0.92 & 100.00 & top100 & 0.79 & 100.00\\
topk & 0.90 & 60.00 & topk & 0.90 & 465.00\\
\bottomrule
\end{tabular}%
}
\caption{Comparison of the fixed-size set predictors and the top-K predictor with $K$ learned based on the validation set. The table reports the mean values of coverage and the average size of answer sets over 15 trials. Results are based on the RESCAL model applied to the WN18 and FB15k datasets.}
\label{tab:fixsize}
\end{table}

\section{Detailed Experimental Setting}\label{app:kge_models}
\subsection{Information About KGE Models and Benchmark Datasets}
We provide the statistics of the benchmark datasets in Table \ref{tab:dataset} and the scoring functions of KGE methods in Table \ref{tab:scoring_functions}.
\begin{table}[t!]
\resizebox{.48\textwidth}{!}{%
\begin{tabular}{@{}llllll@{}}
\toprule
          & \#Entity & \#Relation & \#Training & \#Validation & \#Test \\ \midrule
WN18      & 40,943   & 18       & 141,442    & 5,000        & 5,000  \\
WN18RR    & 40,943   & 11       & 86,835     & 3,034        & 3,134  \\
FB15k     & 14,951   & 1,345    & 483,142    & 50,000       & 59,071 \\
FB15k-237 & 14,541   & 237      & 272,115    & 17,535       & 20,466 \\ \bottomrule
\end{tabular}%
}
\caption{Statistics of benchmark datasets for link prediction task.}
\label{tab:dataset}
\end{table}

\begin{table}[t!]
\centering
\resizebox{.48\textwidth}{!}{%
\begin{tabular}{lc}
\toprule
         & Scoring Function $s(<h,r,t>)$\\ \midrule
TransE \cite{bordes2013translating}   & $-||\mathbf{h}+\mathbf{r}-\mathbf{t}||_{1/2}$\\
RotatE \cite{sun2019rotate}  & $-||\mathbf{h}\circ\mathbf{r}-\mathbf{t}||_{p}$\\
RESCAL \cite{nickel2011rescal}   & $\mathbf{h}^T\mathbf{M}_r\mathbf{t}$\\
DistMult \cite{yang2015distmult} & $\mathbf{h}^Tdiag(\mathbf{r})\mathbf{t}$\\
ComplEx \cite{trouillon2016complex}  & $Re(\mathbf{h}^Tdiag(\mathbf{r})\overline{\mathbf{t}})$\\
ConvE \cite{dettmers2018convolutional}   & $f(vec(f([\overline{\mathbf{h}};\overline{\mathbf{r}}]*\omega))\mathbf{W})\mathbf{t}$\\ \bottomrule
\end{tabular}%
}
\caption{The scoring function $s(<\boldsymbol{h,r,t}>)$ of KGE models used in this paper, where $\boldsymbol{h,r,t}$ denote the embeddings of $h,r,t$, $\circ$ denotes Hadamard product. $\overline{\cdot}$ refers to conjugate for complex vectors in ComplEx, and 2D reshaping for real vectors in ConvE. $*$ is operator for 2D convolution. $\omega$ is the filters and $W$ is the parameters for 2D convolutional layer.}
\label{tab:scoring_functions}
\end{table}

\subsection{Personal Identification Issue in FB15k and FB15k237}
While FB15k and FB15k237 contain information about individuals, it typically focuses on well-known public figures such as celebrities, politicians, and historical figures. Since this information is already widely available online and in various public sources, its inclusion in Freebase doesn't significantly compromise individual privacy compared to datasets containing sensitive personal information.

\subsection{Details of Pre-training KGE Models}
For training KGE models, we use the implementation of LibKGE \cite{libkge} and basically follow the hyperparameter search strategy in \cite{ruffinelli2019you}.
All experiments were conducted on a Linux machine with a 40GB NVIDIA A100 SXM4 GPU.

We first conduct quasi-random hyperparameter search via a Sobol sequence, which aims to distribute hyperparameter settings evenly to avoid "clumping" effects \cite{bergstra2012random}. More specifically, for each dataset
and model, we generated 30 different configurations per valid combination of training type and loss function. we added a short Bayesian optimization
phase (best configuration so far + 30 new trials) to tune the hyperparameters further. All above steps are conducted using Ax framework (\url{https://ax.dev/})

We use a large hyperparameter space including loss functions (pairwise margin ranking with hinge loss, binary cross entropy, cross entropy), regularization techniques (none/L1/L2/L3, dropout), optimizers (Adam, Adagrad), and initialization methods used in the KGE community as hyperparameters. We consider 128, 256, 512 as possible embedding sizes. More details see in \cite[Table 5]{ruffinelli2019you}.

The hyperparameters of the baseline models are located within the software folder we submitted. Concretely, all configuration files (*.yaml) that we use for training baseline models/competing models/models for aggregation can be found in folder "configs".

\section{Calibrated Conformal Predictor}\label{app:cali_cp}
Conformal prediction is a theoretical framework that quantifies predictive uncertainty by ensuring answer sets meet probabilistic guarantees, followed by identifying a nonconformity measure that minimizes the size of these sets. Optimal answer sets are achieved when nonconformity scores accurately reflect the confidence of the predictions.

While calibrating plausibility scores from KGE models should theoretically improve the naive predictor and the nonconformity measure for conformal predictor, our results suggest otherwise. As shown in Table \ref{tab:main1_filter}, \ref{tab:main2_filter}, \ref{tab:main1} and \ref{tab:main2}, Platt predictors yield excessively large answer sets. Further experiments comparing softmax conformal prediction before and after temperature scaling (Table \ref{tab:cali_main}) reveal that temperature scaling generally increases answer set sizes. Although smaller sets are observed for TransE, they are still not competitive with the best predictors for TransE in Table \ref{tab:main1_filter}. These observations contradict our expectations. We next explore the reasons for these outcomes in KGE models.

First, as detailed in Appendix \ref{app:platt}, temperature scaling adjusts plausibility scores by dividing by a temperature parameter $T$, optimized by minimizing negative log-likelihood on validation set. This calibration assumes two key points: (1) \textbf{uniform miscalibration}, where plausibility scores are consistently miscalibrated across the model (e.g., the KGE model is uniformly overconfident or underconfident for all queries); and (2) \textbf{monotonic calibration}, where the relative ordering of plausibility scores aligns with calibrated probabilities. These assumptions are overly stringent for KGE models, which tend to be overconfident with queries that have many correct answers and underconfident with those having fewer correct answers. Additionally, the relative ordering of plausibility scores is highly sensitive to minor hyperparameter changes.

Moreover, applying temperature scaling or other calibration techniques requires formulating link prediction as a classification task. However, the validation set exhibits a long-tail distribution in the number of triples associated with certain entities, i.e. many entities have few or even no associated triples. It leads to insufficient data for effective calibration for entities associated with fewer triples. 

\begin{table}[h!]
\resizebox{\linewidth}{!}{%
\begin{tabular}{cc|cc|cc}
\toprule[1.5pt]
& & \multicolumn{2}{c}{WN18} & \multicolumn{2}{c}{FB15k} \\
model & method & coverage & size & coverage & size \\\midrule
\multirow{2}{*}{TransE} & Softmax & 0.90 & 112.8 & 0.90 & 414.9 \\
                        & Cali & 0.90 & \textbf{63.4} & 0.90 & \textbf{129.0} \\\midrule
\multirow{2}{*}{RotatE} & Softmax & 0.90 & \textbf{1.9} & 0.90 & \textbf{140.4} \\
                        & Cali & 0.90 & 17.4 & 0.90 & 150.6 \\\midrule
\multirow{2}{*}{RESCAL} & Softmax & 0.90 & \textbf{2.1} & 0.90 & \textbf{72.6} \\
                        & Cali & 0.91 & 247.3 & 0.90 & 209.5 \\\midrule
\multirow{2}{*}{DistMult} & Softmax & 0.90 & \textbf{2.0} & 0.90 & \textbf{28.4} \\
                        & Cali & 0.90 & 26.8 & 0.90 & 240.5 \\\midrule
\multirow{2}{*}{ComplEx} & Softmax & 0.90 & \textbf{1.1} & 0.90 & \textbf{37.2} \\
                        & Cali & 0.90 & 18.2 & 0.90 & 173.6 \\\midrule
\multirow{2}{*}{ConvE} & Softmax & 0.90 & \textbf{1.6} & 0.90 & \textbf{44.4} \\
                        & Cali & 0.90 & 61.2 & 0.90 & 177.7 \\
\bottomrule
\end{tabular}%
}
\caption{Comparison of the filtered answer sets between the Softmax conformal predictor (Softmax) and the conformal predictor with temperature scaling applied to the Softmax predictor (Cali). The best predictors, which meet the coverage desideratum and minimize answer set size, are highlighted in bold.}
\label{tab:cali_main}
\end{table}

\section{Further Results for Coverage \& Set Size Evaluation}\label{app:cover}
\subsection{Coverage and Set Size on WN18RR and FB15k237}
We repeated the experiment on WN18RR and FB15k237, datasets known to be more challenging than WN18 and FB15k due to the removal of inverse relations \cite{toutanova2015observed, dettmers2018convolutional}. 

The results for the filtered answer sets are presented in Table \ref{tab:main2_filter}, while the unfiltered results are available in Table \ref{tab:main2} in Appendix \ref{app:cover}. The conclusions from Experiment 1 remain consistent; however, we observe a significant increase in set sizes for all set predictors, particularly for WN18RR. This increase is desirable, as it aligns with the adaptiveness desideratum, where the set predictor is expected to output smaller sets for simple queries and larger sets for harder ones.

\begin{table*}[]
\resizebox{\textwidth}{!}{%
\begin{tabular}{ccccc|ccccc}
\toprule[2pt]
\multicolumn{5}{c}{WN18RR} & \multicolumn{5}{c}{FB15k237} \\
model & MR & methods & coverage & size & model & MR & methods & coverage & size \\\midrule[1.5pt]

 &  & naive & 0.92 (0.002) & 12592.31 (39.396) &  &  & naive & 0.90 (0.006) & 805.70 (38.319) \\
&  & Platt & 0.90 (0.002) & 10921.01 (32.441) &  &  & Platt & 0.90 (0.006) & 832.14 (36.711) \\
 &  & topk & 0.90 (0.002) & \textbf{3571.51 (144.178)} &  &  & topk & 0.90 (0.001) & 875.53 (7.835) \\
 &  & {\ul NegScore} & 0.90 (0.002) & 9409.77 (252.614) &  &  & {\ul NegScore} & 0.90 (0.001) & 1367.25 (39.240) \\
 &  & {\ul Softmax} & 0.90 (0.001) & 4864.10 (160.461) &  &  & {\ul Softmax} & 0.90 (0.001) & \textbf{340.67 (3.099)} \\
\multirow{-6}{*}{TransE} & \multirow{-6}{*}{1849.47 (20.933)} & {\ul Minmax} & 0.90 (0.001) & 4371.36 (172.089) & \multirow{-6}{*}{TransE} & \multirow{-6}{*}{206.62 (2.105)} & {\ul Minmax} & 0.90 (0.001) & 482.96 (8.236) \\\midrule

 &  & naive & 0.98 (0.003) & 29054.22 (78.389) &  &  & naive & 0.99 (0.000) & 4564.86 (16.479) \\
 &  & Platt & 0.92 (0.003) & 23041.24 (67.999) &  &  & Platt & 0.95 (0.001) & 1851.22 (12.442) \\
 &  & topk & 0.90 (0.003) & \textbf{7780.12 (1372.505)} &  &  & topk & 0.90 (0.001) & 786.96 (9.659) \\
 &  & {\ul NegScore} & 0.90 (0.004) & 10135.01 (887.572) &  &  & {\ul NegScore} & 0.90 (0.001) & 396.30 (6.841) \\
 &  & {\ul Softmax} & 0.90 (0.003) & 8469.82 (1332.540) &  &  & {\ul Softmax} & 0.90 (0.001) & \textbf{309.38 (4.527)} \\
\multirow{-6}{*}{RotatE} & \multirow{-6}{*}{2402.47 (226.057)} & {\ul Minmax} & 0.90 (0.003) & 8026.50 (691.561) & \multirow{-6}{*}{RotatE} & \multirow{-6}{*}{167.92 (3.340)} & {\ul Minmax} & 0.90 (0.001) & 310.32 (5.611) \\\midrule

 &  & naive & {\color[HTML]{FF0000} 0.82 (0.004)} & 19604.12 (54.324) &  &  & naive & {\color[HTML]{FF0000} 0.75 (0.026)} & 311.80 (59.631) \\
 &  & Platt & 0.91 (0.006) & 25156.82 (67.922) &  &  & Platt & {\color[HTML]{FF0000} 0.85 (0.017)} & 492.83 (68.264) \\
 &  & topk & 0.90 (0.006) & 20571.25 (619.329) &  &  & topk & 0.90 (0.001) & 810.16 (8.750) \\
 &  & {\ul NegScore} & 0.90 (0.006) & 19813.44 (476.890) &  &  & {\ul NegScore} & 0.90 (0.001) & 581.74 (16.583) \\
 &  & {\ul Softmax} & 0.90 (0.006) & 25146.85 (397.278) &  &  & {\ul Softmax} & 0.90 (0.001) & \textbf{261.58 (5.352)} \\
\multirow{-6}{*}{RESCAL} & \multirow{-6}{*}{5080.82 (157.027)} & {\ul Minmax} & 0.90 (0.005) & \textbf{18262.03 (484.686)} & \multirow{-6}{*}{RESCAL} & \multirow{-6}{*}{197.71 (7.228)} & {\ul Minmax} & 0.90 (0.001) & 356.23 (20.744) \\\midrule

 &  & naive & {\color[HTML]{FF0000} 0.87 (0.014)} & 22687.34 (1040.595) &  &  & naive & {\color[HTML]{FF0000} 0.82 (0.008)} & 852.93 (110.675) \\
 &  & Platt & 0.90 (0.005) & 26100.71 (766.341) &  &  & Platt & {\color[HTML]{FF0000} 0.88 (0.007)} & 1236.71 (122.667) \\
 &  & topk & 0.90 (0.005) & 18220.44 (660.313) &  &  & topk & 0.90 (0.001) & 785.39 (7.479) \\
 &  & {\ul NegScore} & 0.90 (0.005) & 22735.97 (843.241) &  &  & {\ul NegScore} & 0.90 (0.001) & 340.43 (8.278) \\
 &  & {\ul Softmax} & 0.90 (0.004) & 24347.85 (1756.093) &  &  & {\ul Softmax} & 0.90 (0.001) & \textbf{276.85 (4.932)} \\
\multirow{-6}{*}{DistMult} & \multirow{-6}{*}{4325.85 (153.189)} & {\ul Minmax} & 0.90 (0.005) & \textbf{17555.05 (822.261)} & \multirow{-6}{*}{DistMult} & \multirow{-6}{*}{194.19 (4.581)} & {\ul Minmax} & 0.90 (0.001) & 352.42 (31.161) \\\midrule

 &  & naive & {\color[HTML]{FF0000} 0.45 (0.008)} & 5939.57 (192.280) &  &  & naive & {\color[HTML]{FF0000} 0.85 (0.010)} & 1027.42 (134.273) \\
 &  & Platt & {\color[HTML]{FF0000} 0.83 (0.008)} & 15307.46 (374.11) &  &  & Platt & 0.92 (0.010) & 1701.31 (145.989) \\
 &  & topk & 0.90 (0.006) & 19785.19 (410.275) &  &  & topk & 0.90 (0.001) & 757.90 (5.512) \\
 &  & {\ul NegScore} & 0.90 (0.007) & 19858.11 (221.472) &  &  & {\ul NegScore} & 0.90 (0.001) & 319.42 (6.025) \\
 &  & {\ul Softmax} & 0.90 (0.004) & 18194.32 (595.990) &  &  & {\ul Softmax} & 0.90 (0.001) & \textbf{271.05 (2.988)} \\
\multirow{-6}{*}{ComplEx} & \multirow{-6}{*}{4117.56 (127.304)} & {\ul Minmax} & 0.90 (0.003) & \textbf{14101.85 (447.917)} & \multirow{-6}{*}{ComplEx} & \multirow{-6}{*}{183.58 (3.182)} & {\ul Minmax} & 0.90 (0.001) & 365.28 (72.147) \\\midrule

 &  & naive & {\color[HTML]{FF0000} 0.25 (0.006)} & 1131.72 (88.750) &  &  & naive & 0.95 (0.006) & 1072.99 (100.842) \\
 &  & Platt & {\color[HTML]{FF0000} 0.82 (0.006)} & 10955.50 (800.116) &  &  & Platt & 0.94 (0.005) & 814.70 (75.111) \\
 &  & topk & 0.90 (0.005) & 18270.13 (1047.722) &  &  & topk & 0.90 (0.001) & 752.37 (5.652) \\
 &  & {\ul NegScore} & 0.90 (0.004) & 21094.54 (687.705) &  &  & {\ul NegScore} & 0.90 (0.001) & 718.24 (44.095) \\
 &  & {\ul Softmax} & 0.93 (0.006) & 19851.28 (756.774) &  &  & {\ul Softmax} & 0.90 (0.001) & 270.40 (3.336) \\
\multirow{-6}{*}{ConvE} & \multirow{-6}{*}{4635.63 (151.271)} & {\ul Minmax} & 0.90 (0.003) & \textbf{17400.91 (644.163)} & \multirow{-6}{*}{ConvE} & \multirow{-6}{*}{185.19 (1.636)} & {\ul Minmax} & 0.90 (0.001) & \textbf{242.64 (1.843)}\\\bottomrule[2pt]
\end{tabular}%
}
\caption{Quality of the filtered answer sets on WN18RR and FB15k237 datasets. This table presents the mean rank (MR) of KGE models, along with the coverage and size of answer sets generated using various set predictors. Conformal predictors are underlined. Means and standard deviations over 15 trials are reported at the $10\%$ level ($\epsilon=0.1$). Predictors that fail to meet the coverage threshold of $1-\epsilon$ (0.9) are highlighted in red. The best predictors, which meet the coverage desideratum and minimize answer set size, are highlighted in bold.}
\label{tab:main2_filter}
\end{table*}

\begin{table*}[]
\resizebox{\textwidth}{!}{%
\begin{tabular}{ccccc|ccccc}
\toprule[2pt]
\multicolumn{5}{c}{WN18} & \multicolumn{5}{c}{FB15k} \\
model & MR & methods & coverage & size & model & MR & methods & coverage & size \\\midrule[1.5pt]

 &  & naive & {\color[HTML]{FF0000} 0.44 (0.004)} & 26.39 (1.278) &  &  & naive & {\color[HTML]{FF0000} 0.73 (0.001)} & 406.43 (0.827) \\
 &  & Platt & {\color[HTML]{FF0000} 0.85(0.002)} & 4060.21 (89.563) &  &  & Platt & {\color[HTML]{FF0000} 0.84 (0.002)} & 1337.54 (44.123) \\
 &  & topk & 0.90 (0.000) & 54.67 (0.789) &  &  & topk & 0.90 (0.000) & 401.00 (1.461) \\
 &  & {\ul NegScore} & 0.90 (0.001) & \textbf{37.29 (0.721)} &  &  & {\ul NegScore} & 0.90 (0.001) & \textbf{155.40 (0.623)} \\
 &  & {\ul Softmax} & 0.90 (0.001) & 129.67 (4.650) &  &  & {\ul Softmax} & 0.90 (0.000) & 569.80 (2.423) \\
\multirow{-6}{*}{TransE} & \multirow{-6}{*}{261.27 (6.365)} & {\ul Minmax} & 0.90 (0.001) & 130.45 (5.100) & \multirow{-6}{*}{TransE} & \multirow{-6}{*}{188.23 (0.187)} & {\ul Minmax} & 0.90 (0.000) & 433.99 (3.242) \\\midrule

 &  & naive & 0.91 (0.003) & 17704.98 (117.899) &  &  & naive & {\color[HTML]{FF0000} 0.71 (0.003)} & 889.73 (5.109) \\
 &  & Platt & 0.90 (0.002) & 16966.23 (116.434) &  &  & Platt & {\color[HTML]{FF0000} 0.88 (0.002)} & 1292.32 (7.532) \\
 &  & topk & 0.90 (0.001) & 57.73 (2.594) &  &  & topk & 0.90 (0.000) & 479.50 (4.066) \\
 &  & {\ul NegScore} & 0.90 (0.002) & 17.87 (0.022) &  &  & {\ul NegScore} & 0.90 (0.001) & 209.36 (0.665) \\
 &  & {\ul Softmax} & 0.90 (0.001) & \textbf{13.61 (0.401)} &  &  & {\ul Softmax} & 0.90 (0.000) & 277.84 (5.434) \\
\multirow{-6}{*}{RotatE} & \multirow{-6}{*}{493.17 (44.041)} & {\ul Minmax} & 0.90 (0.003) & 20.53 (0.693) & \multirow{-6}{*}{RotatE} & \multirow{-6}{*}{212.70 (1.003)} & {\ul Minmax} & 0.90 (0.001) & \textbf{198.67 (1.117)} \\\midrule

 &  & naive & {\color[HTML]{FF0000} 0.58 (0.011)} & 316.68 (24.776) &  &  & naive & {\color[HTML]{FF0000} 0.58 (0.016)} & 274.69 (13.802) \\
 &  & Platt & {\color[HTML]{FF0000} 0.80 (0.005)} & 2035.13 (222.193) &  &  & Platt & {\color[HTML]{FF0000} 0.87 (0.003)} & 755.23 (22.263) \\
 &  & topk & 0.90 (0.001) & 61.60 (1.744) &  &  & topk & 0.90 (0.000) & 464.00 (2.556) \\
 &  & {\ul NegScore} & 0.91 (0.001) & 55.36 (3.529) &  &  & {\ul NegScore} & 0.90 (0.001) & 327.14 (4.569) \\
 &  & {\ul Softmax} & 0.90 (0.001) & \textbf{15.16 (0.358)} &  &  & {\ul Softmax} & 0.90 (0.000) & \textbf{183.69 (3.650)} \\
\multirow{-6}{*}{RESCAL} & \multirow{-6}{*}{338.24 (21.476)} & {\ul Minmax} & 0.90 (0.002) & 18.81 (0.080) & \multirow{-6}{*}{RESCAL} & \multirow{-6}{*}{215.57 (1.548)} & {\ul Minmax} & 0.90 (0.001) & 237.00 (5.822) \\\midrule

 &  & naive & {\color[HTML]{FF0000} 0.47 (0.002)} & 51.35 (7.023) &  &  & naive & {\color[HTML]{FF0000} 0.35 (0.013)} & 162.65 (1.103) \\
 &  & Platt & {\color[HTML]{FF0000} 0.84 (0.001)} & 1282.45 (204.44) &  &  & Platt & 0.90 (0.001) & 625.33 (2.300) \\
 &  & topk & 0.90 (0.001) & 64.80 (2.587) &  &  & topk & 0.90 (0.000) & 430.00 (1.549) \\
 &  & {\ul NegScore} & 0.90 (0.001) & 2261.78 (405.035) &  &  & {\ul NegScore} & 0.90 (0.000) & 210.35 (2.876) \\
 &  & {\ul Softmax} & 0.90 (0.001) & \textbf{9.16 (0.337)} &  &  & {\ul Softmax} & 0.90 (0.000) & \textbf{143.23 (4.601)} \\
\multirow{-6}{*}{DistMult} & \multirow{-6}{*}{386.83 (20.312)} & {\ul Minmax} & 0.90 (0.002) & 18.04 (0.525) & \multirow{-6}{*}{DistMult} & \multirow{-6}{*}{196.36 (0.746)} & {\ul Minmax} & 0.90 (0.000) & 197.79 (0.892) \\\midrule

 &  & naive & 0.94 (0.007) & 19984.22 (142.118) &  &  & naive & {\color[HTML]{FF0000} 0.28 (0.011)} & 189.44 (1.424) \\
 &  & Platt & {\color[HTML]{FF0000} 0.86 (0.002)} & 16802.14 (125.322) &  &  & Platt & 0.90 (0.001) & 1082.11 (9.211) \\
 &  & topk & 0.91 (0.002) & 46.73 (1.948) &  &  & topk & 0.90 (0.000) & 485.60 (4.800) \\
 &  & {\ul NegScore} & 0.90 (0.003) & 18.36 (0.008) &  &  & {\ul NegScore} & 0.90 (0.001) & 257.63 (9.161) \\
 &  & {\ul Softmax} & 0.90 (0.001) & \textbf{7.63 (0.133)} &  &  & {\ul Softmax} & 0.90 (0.001) & \textbf{178.62 (3.561)} \\
\multirow{-6}{*}{ComplEx} & \multirow{-6}{*}{467.12 (27.864)} & {\ul Minmax} & 0.90 (0.002) & 19.72 (0.172) & \multirow{-6}{*}{ComplEx} & \multirow{-6}{*}{216.65 (1.698)} & {\ul Minmax} & 0.90 (0.001) & 278.10 (20.531) \\\midrule

 &  & naive & {\color[HTML]{FF0000} 0.50 (0.015)} & 18.32 (1.536) &  &  & naive & {\color[HTML]{FF0000} 0.45 (0.026)} & 932.71 (112.137) \\
 &  & Platt & {\color[HTML]{FF0000} 0.84 (0.007)} & 95.01 (1.772) &  &  & Platt & {\color[HTML]{FF0000} 0.88 (0.014)} & 4984.11 (699.111) \\
 &  & topk & 0.90 (0.001) & 60.27 (1.569) &  &  & topk & 0.90 (0.000) & 455.13 (3.384) \\
 &  & {\ul NegScore} & 0.90 (0.002) & 17.47 (0.617) &  &  & {\ul NegScore} & 0.90 (0.000) & \textbf{203.47 (2.291)} \\
 &  & {\ul Softmax} & 0.90 (0.001) & \textbf{8.45 (0.666)} &  &  & {\ul Softmax} & 0.90 (0.000) & 217.82 (6.516) \\
\multirow{-6}{*}{ConvE} & \multirow{-6}{*}{327.52 (12.602)} & {\ul Minmax} & 0.90 (0.001) & 17.80 (0.161) & \multirow{-6}{*}{ConvE} & \multirow{-6}{*}{216.37 (1.848)} & {\ul Minmax} & 0.90 (0.000) & 206.62 (8.665)\\
\bottomrule[2pt]
\end{tabular}%
}
\caption{Quality of the answer sets on WN18 and FB15k datasets. This table presents the mean rank (MR) of KGE models, along with the coverage and size of answer sets generated using various set predictors. Conformal predictors are underlined. Means and standard deviations over 15 trials are reported at the $10\%$ level ($\epsilon=0.1$). Predictors that fail to meet the coverage threshold of $1-\epsilon$ (0.9) are highlighted in red. The best predictors, which meet the coverage desideratum and minimize answer set size, are highlighted in bold.}
\label{tab:main1}
\end{table*}

\begin{table*}[]
\resizebox{\textwidth}{!}{%
\begin{tabular}{ccccc|ccccc}
\toprule[2pt]
\multicolumn{5}{c}{WN18RR} & \multicolumn{5}{c}{FB15k237} \\
model & MR & methods & coverage & size & model & MR & methods & \multicolumn{1}{c}{coverage} & \multicolumn{1}{c}{size} \\\midrule[1.5pt]

 &  & naive & {\color[HTML]{FF0000} 0.92 (0.002)} & 12606.42 (39.399) &  &  & naive & {\color[HTML]{FF0000} 0.90 (0.006)} & 997.60 (40.772) \\
 &  & Platt & 0.90 (0.002) & 10935.81 (31.241) &  &  & Platt & 0.90 (0.006) & 1022.32 (35.123) \\
 &  & topk & 0.90 (0.002) & \textbf{3585.87 (144.179)} &  &  & topk & 0.90 (0.001) & 993.33 (8.252) \\
 &  & {\ul NegScore} & 0.90 (0.002) & 9424.01 (252.614) &  &  & {\ul NegScore} & 0.90 (0.001) & 1529.21 (38.853) \\
 &  & {\ul Softmax} & 0.90 (0.001) & 4878.40 (160.461) &  &  & {\ul Softmax} & 0.90 (0.001) & \textbf{480.91 (4.252)} \\
\multirow{-6}{*}{TransE} & \multirow{-6}{*}{1863.55 (20.933)} & {\ul Minmax} & 0.90 (0.001) & 4385.71 (172.090) & \multirow{-6}{*}{TransE} & \multirow{-6}{*}{378.52 (1.570)} & {\ul Minmax} & 0.90 (0.001) & 608.92 (8.100) \\\midrule

 &  & naive & 0.98 (0.003) & 29068.66 (78.388) &  &  & naive & {\color[HTML]{FF0000} 0.99 (0.000)} & 4776.32 (16.527) \\
 &  & Platt & 0.92 (0.003) & 23065.33 (65.139) &  &  & Platt & 0.95 (0.001) & 1992.54 (12.111) \\
 &  & topk & 0.90 (0.003) & \textbf{7794.53 (1372.506)} &  &  & topk & 0.90 (0.001) & 899.20 (10.186) \\
 &  & {\ul NegScore} & 0.90 (0.004) & 10149.42 (887.574) &  &  & {\ul NegScore} & 0.90 (0.001) & 601.35 (6.779) \\
 &  & {\ul Softmax} & 0.90 (0.003) & 8484.13 (1332.541) &  &  & {\ul Softmax} & 0.90 (0.001) & \textbf{420.48 (5.493)} \\
\multirow{-6}{*}{RotatE} & \multirow{-6}{*}{2416.60 (226.054)} & {\ul Minmax} & 0.90 (0.003) & 8040.94 (691.558) & \multirow{-6}{*}{RotatE} & \multirow{-6}{*}{334.72 (2.449)} & {\ul Minmax} & 0.90 (0.001) & 494.12 (7.390) \\\midrule

 &  & naive & {\color[HTML]{FF0000} 0.82 (0.004)} & 19617.55 (54.247) &  &  & naive & {\color[HTML]{FF0000} 0.75 (0.026)} & 518.49 (60.728) \\
 &  & Platt & 0.91 (0.006) & 25178.84 (67.922) &  &  & Platt & {\color[HTML]{FF0000} 0.85 (0.017)} & 634.23 (67.233) \\
 &  & topk & 0.90 (0.006) & 20585.67 (619.329) &  &  & topk & 0.90 (0.001) & 923.93 (9.205) \\
 &  & {\ul NegScore} & 0.90 (0.006) & 19827.86 (476.890) &  &  & {\ul NegScore} & 0.90 (0.001) & 794.21 (16.356) \\
 &  & {\ul Softmax} & 0.90 (0.006) & 25161.15 (397.282) &  &  & {\ul Softmax} & 0.90 (0.001) & \textbf{441.73 (2.242)} \\
\multirow{-6}{*}{RESCAL} & \multirow{-6}{*}{5095.01 (157.027)} & {\ul Minmax} & 0.90 (0.005) & \textbf{18276.47 (484.688)} & \multirow{-6}{*}{RESCAL} & \multirow{-6}{*}{361.48 (5.994)} & {\ul Minmax} & 0.90 (0.001) & 550.94 (20.262) \\\midrule

 &  & naive & {\color[HTML]{FF0000} 0.87 (0.014)} & 22700.68 (1040.718) &  &  & naive & {\color[HTML]{FF0000} 0.82 (0.008)} & 1064.07 (111.022) \\
 &  & Platt & 0.90 (0.005) & 26117.52 (764.112) &  &  & Platt & {\color[HTML]{FF0000} 0.88 (0.007)} & 1385.57 (122.190) \\
 &  & topk & 0.90 (0.005) & 18234.80 (660.315) &  &  & topk & 0.90 (0.001) & 895.13 (7.847) \\
 &  & {\ul NegScore} & 0.90 (0.005) & 22750.38 (843.239) &  &  & {\ul NegScore} & 0.90 (0.001) & 446.12 (7.461) \\
 &  & {\ul Softmax} & 0.90 (0.004) & 24362.11 (1756.095) &  &  & {\ul Softmax} & 0.90 (0.001) & \textbf{429.23 (4.660)} \\
\multirow{-6}{*}{DistMult} & \multirow{-6}{*}{4340.04 (153.190)} & {\ul Minmax} & 0.90 (0.005) & \textbf{17569.49 (822.261)} & \multirow{-6}{*}{DistMult} & \multirow{-6}{*}{343.41 (4.267)} & {\ul Minmax} & 0.90 (0.001) & 553.92 (30.822) \\\midrule

 &  & naive & {\color[HTML]{FF0000} 0.45 (0.008)} & \textbf{5950.18 (192.118)} &  &  & naive & {\color[HTML]{FF0000} 0.85 (0.010)} & 1239.13 (134.671) \\
 &  & Platt & {\color[HTML]{FF0000} 0.83 (0.008)} & 15325.22 (372.89) &  &  & Platt & 0.92 (0.010) & 1852.47 (142.284) \\
 &  & topk & 0.90 (0.006) & 19799.60 (410.282) &  &  & topk & 0.90 (0.001) & 865.33 (5.839) \\
 &  & {\ul NegScore} & 0.90 (0.007) & 19872.52 (221.472) &  &  & {\ul NegScore} & 0.90 (0.001) & 427.18 (5.201) \\
 &  & {\ul Softmax} & 0.90 (0.004) & 18208.61 (595.990) &  &  & {\ul Softmax} & 0.90 (0.001) & \textbf{404.95 (1.924)} \\
\multirow{-6}{*}{ComplEx} & \multirow{-6}{*}{4131.71 (127.296)} & {\ul Minmax} & 0.90 (0.003) & 14116.23 (447.920) & \multirow{-6}{*}{ComplEx} & \multirow{-6}{*}{329.62 (2.596)} & {\ul Minmax} & 0.90 (0.001) & 564.83 (71.526) \\\midrule

 &  & naive & {\color[HTML]{FF0000} 0.25 (0.006)} & 1144.71 (88.743) &  &  & naive & {\color[HTML]{FF0000} 0.95 (0.006)} & 1286.00 (100.851) \\
 &  & Platt & {\color[HTML]{FF0000} 0.82 (0.006)} & 10973.44 (801.106) &  &  & Platt & 0.94 (0.005) & 964.33 (73.154) \\
 &  & topk & 0.90 (0.005) & 18284.53 (1047.723) &  &  & topk & 0.90 (0.001) & 860.40 (6.141) \\
 &  & {\ul NegScore} & 0.90 (0.004) & 21108.92 (687.713) &  &  & {\ul NegScore} & 0.90 (0.001) & 931.44 (44.128) \\
 &  & {\ul Softmax} & 0.93 (0.006) & 19865.73 (756.774) &  &  & {\ul Softmax} & 0.90 (0.001) & \textbf{369.22 (2.583)} \\
\multirow{-6}{*}{ConvE} & \multirow{-6}{*}{4649.83 (151.263)} & {\ul Minmax} & 0.90 (0.003) & \textbf{17415.35 (644.162)} & \multirow{-6}{*}{ConvE} & \multirow{-6}{*}{339.42 (2.611)} & {\ul Minmax} & 0.90 (0.001) & 441.82 (2.828)\\\bottomrule[2pt]
\end{tabular}%
}
\caption{Quality of the answer sets on WN18RR and FB15k237 datasets. This table presents the mean rank (MR) of KGE models, along with the coverage and size of answer sets generated using various set predictors. Conformal predictors are underlined. Means and standard deviations over 15 trials are reported at the $10\%$ level ($\epsilon=0.1$). Predictors that fail to meet the coverage threshold of $1-\epsilon$ (0.9) are highlighted in red. The best predictors, which meet the coverage desideratum and minimize answer set size, are highlighted in bold.}
\label{tab:main2}
\end{table*}

\section{Further Results for Adaptiveness Evaluation}\label{app:adap}
\begin{figure*}[h!]
    \centering
    \includegraphics[width=\linewidth]{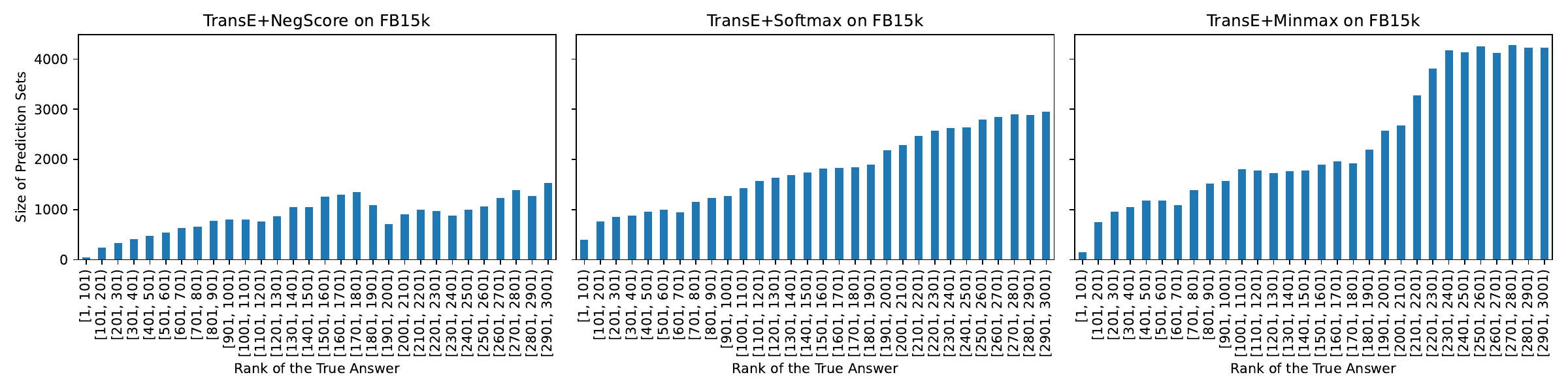}
    \caption{This figure shows the size of answer sets stratified by the difficulty level of queries. It shows the adaptiveness of different conformal predictors (built on TransE models) on the FB15k dataset.}
    \label{fig:adap_TransE_FB15k}
\end{figure*}

\begin{figure*}[h!]
    \centering
    \includegraphics[width=\linewidth]{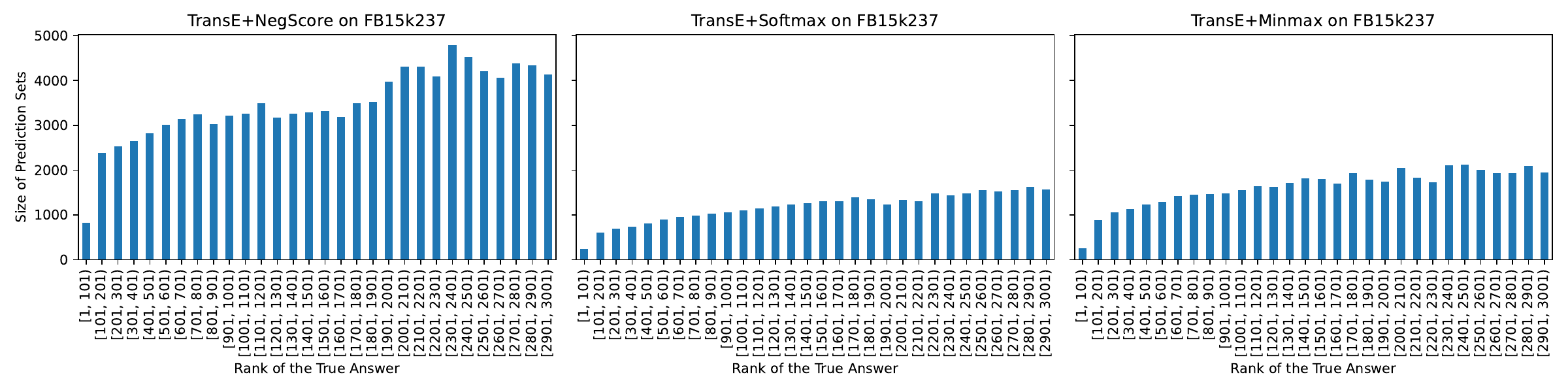}
    \caption{This figure shows the size of answer sets stratified by the difficulty level of queries. It shows the adaptiveness of different conformal predictors (built on TransE models) on the FB15k237 dataset.}
    \label{fig:adap_TransE_FB15k237}
\end{figure*}

\begin{figure*}[h!]
    \centering
    \includegraphics[width=\linewidth]{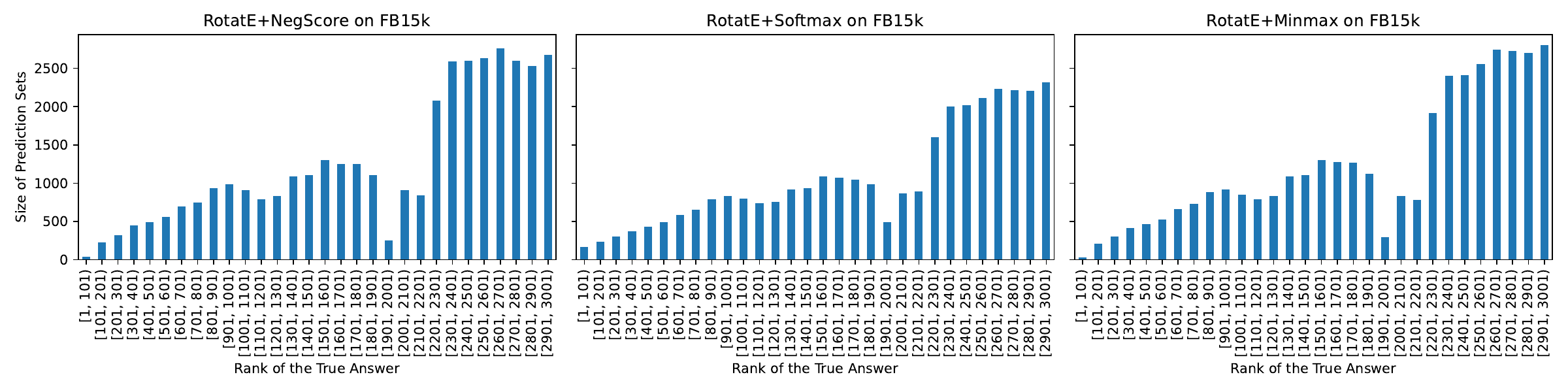}
    \caption{This figure shows the size of answer sets stratified by the difficulty level of queries. It shows the adaptiveness of different conformal predictors (built on RotatE models) on the FB15k dataset.}
    \label{fig:adap_RotatE_FB15k}
\end{figure*}

\begin{figure*}[h!]
    \centering
    \includegraphics[width=\linewidth]{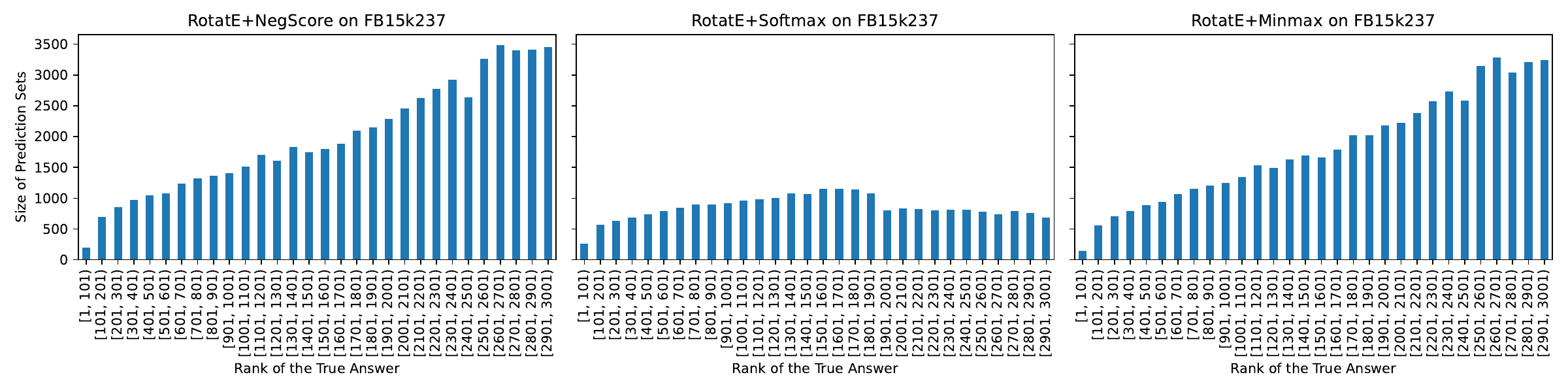}
    \caption{This figure shows the size of answer sets stratified by the difficulty level of queries. It shows the adaptiveness of different conformal predictors (built on RotatE models) on the FB15k237 dataset.}
    \label{fig:adap_RotatE_FB15k237}
\end{figure*}

\begin{figure*}[h!]
    \centering
    \includegraphics[width=\linewidth]{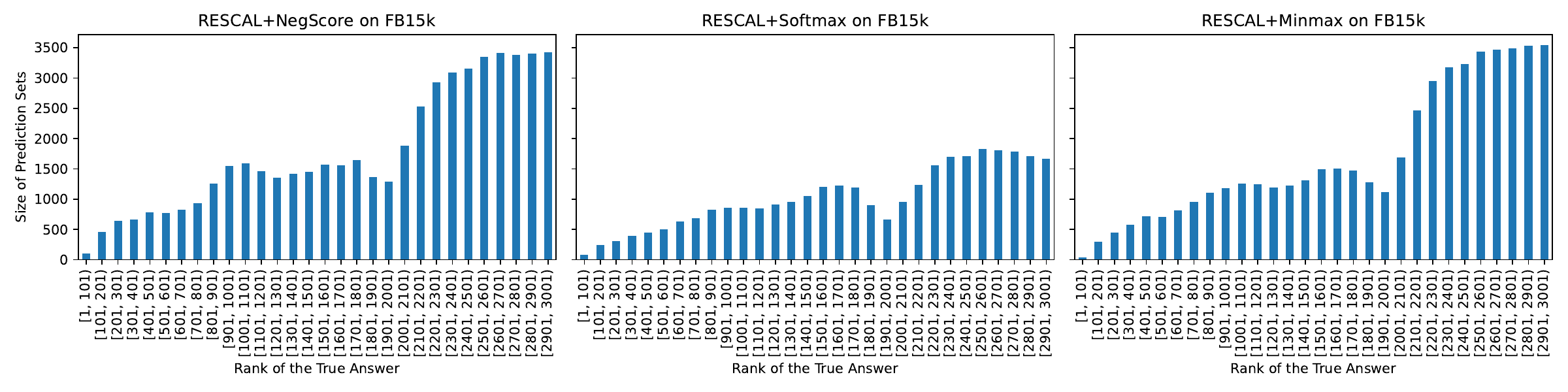}
    \caption{This figure shows the size of answer sets stratified by the difficulty level of queries. It shows the adaptiveness of different conformal predictors (built on RESCAL models) on the FB15k dataset.}
    \label{fig:adap_RESCAL_FB15k}
\end{figure*}

\begin{figure*}[h!]
    \centering
    \includegraphics[width=\linewidth]{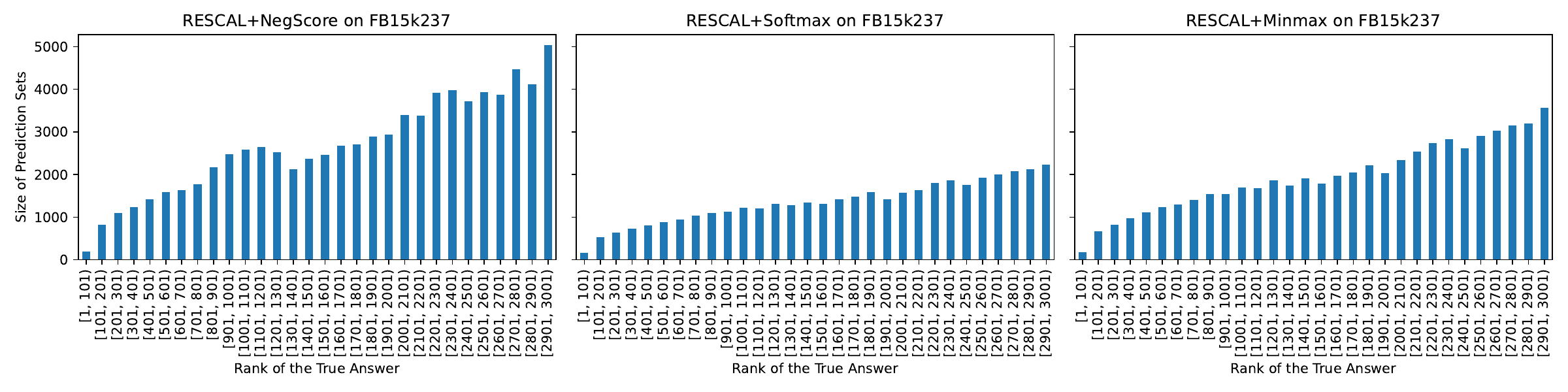}
    \caption{This figure shows the size of answer sets stratified by the difficulty level of queries. It shows the adaptiveness of different conformal predictors (built on RESCAL models) on the FB15k237 dataset.}
    \label{fig:adap_RESCAL_FB15k237}
\end{figure*}

\begin{figure*}[h!]
    \centering
    \includegraphics[width=\linewidth]{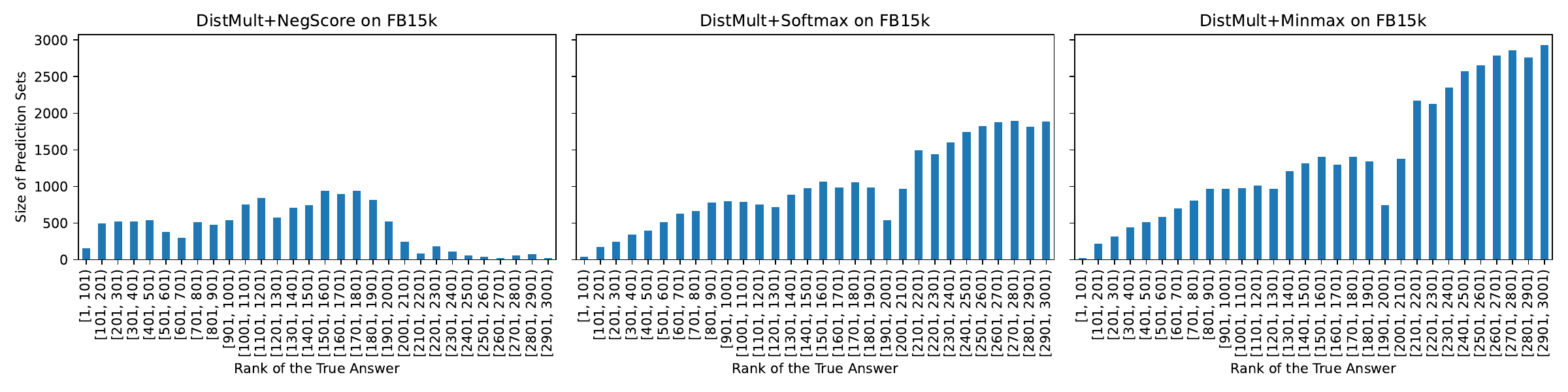}
    \caption{This figure shows the size of answer sets stratified by the difficulty level of queries. It shows the adaptiveness of different conformal predictors (built on DistMult models) on the FB15k dataset.}
    \label{fig:adap_DistMult_FB15k}
\end{figure*}

\begin{figure*}[h!]
    \centering
    \includegraphics[width=\linewidth]{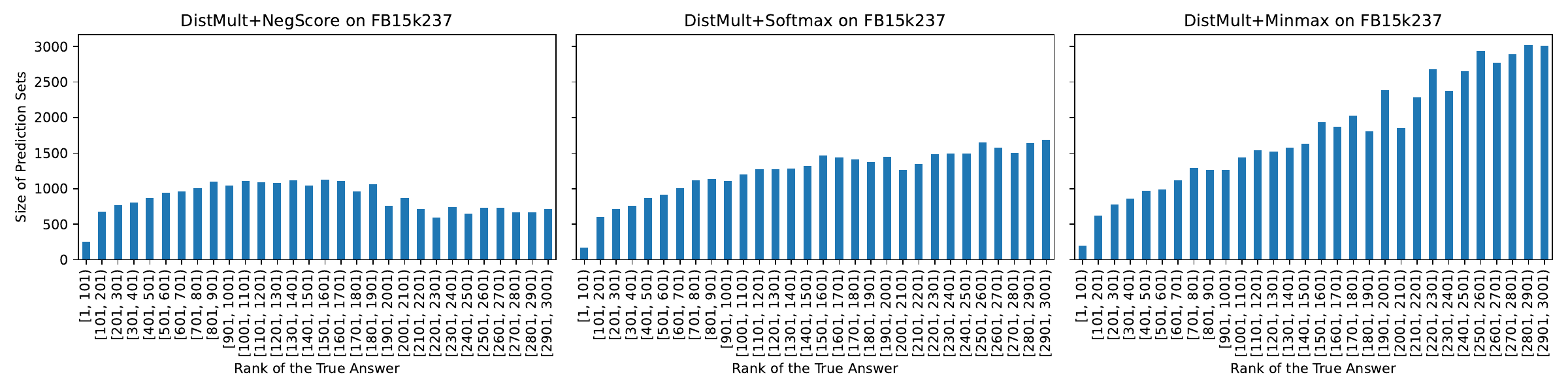}
    \caption{This figure shows the size of answer sets stratified by the difficulty level of queries. It shows the adaptiveness of different conformal predictors (built on DistMult models) on the FB15k237 dataset.}
    \label{fig:adap_DistMult_FB15k237}
\end{figure*}

\begin{figure*}[h!]
    \centering
    \includegraphics[width=\linewidth]{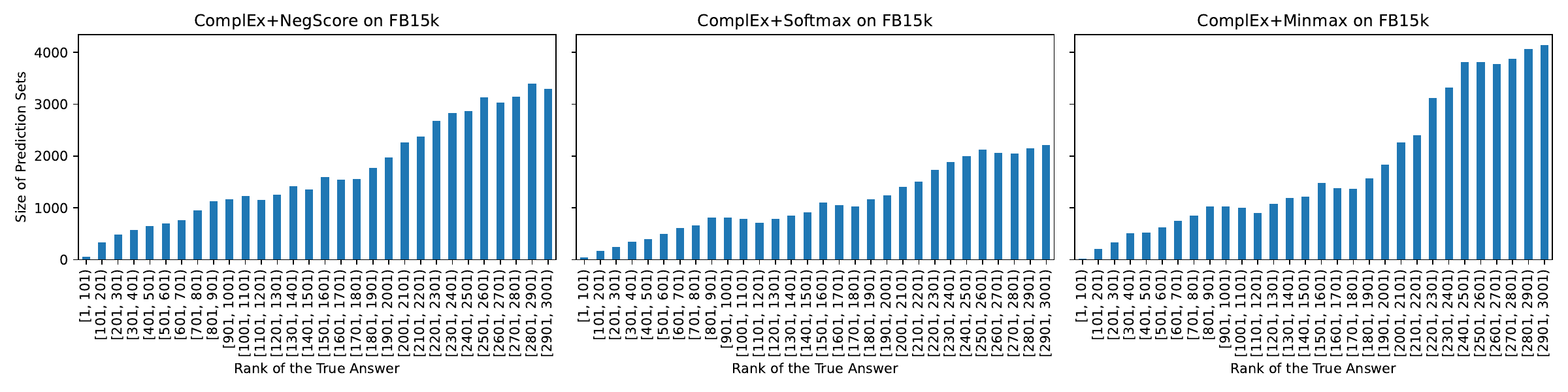}
    \caption{This figure shows the size of answer sets stratified by the difficulty level of queries. It shows the adaptiveness of different conformal predictors (built on ComplEx models) on the FB15k dataset.}
    \label{fig:adap_ComplEx_FB15k}
\end{figure*}

\begin{figure*}[h!]
    \centering
    \includegraphics[width=\linewidth]{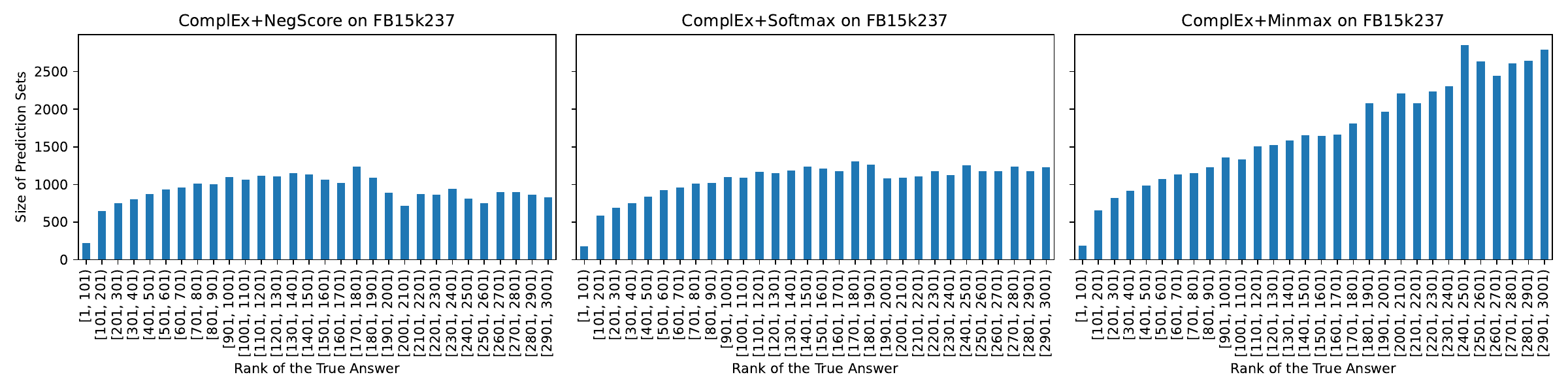}
    \caption{This figure shows the size of answer sets stratified by the difficulty level of queries. It shows the adaptiveness of different conformal predictors (built on ComplEx models) on the FB15k237 dataset.}
    \label{fig:adap_ComplEx_FB15k237}
\end{figure*}

\begin{figure*}[h!]
    \centering
    \includegraphics[width=\linewidth]{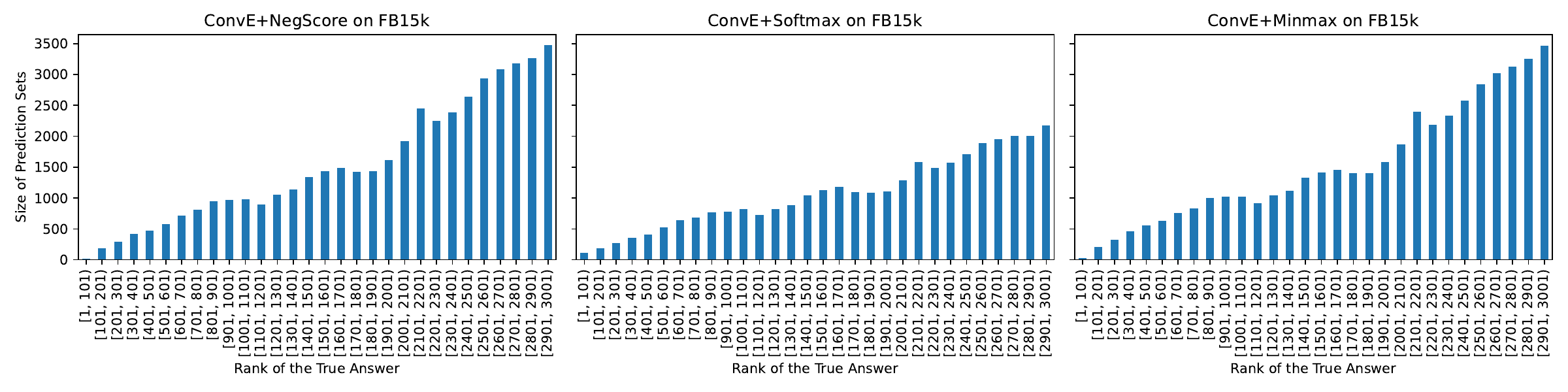}
    \caption{This figure shows the size of answer sets stratified by the difficulty level of queries. It shows the adaptiveness of different conformal predictors (built on ConvE models) on the FB15k dataset.}
    \label{fig:adap_ConvE_FB15k}
\end{figure*}

\begin{figure*}[h!]
    \centering
    \includegraphics[width=\linewidth]{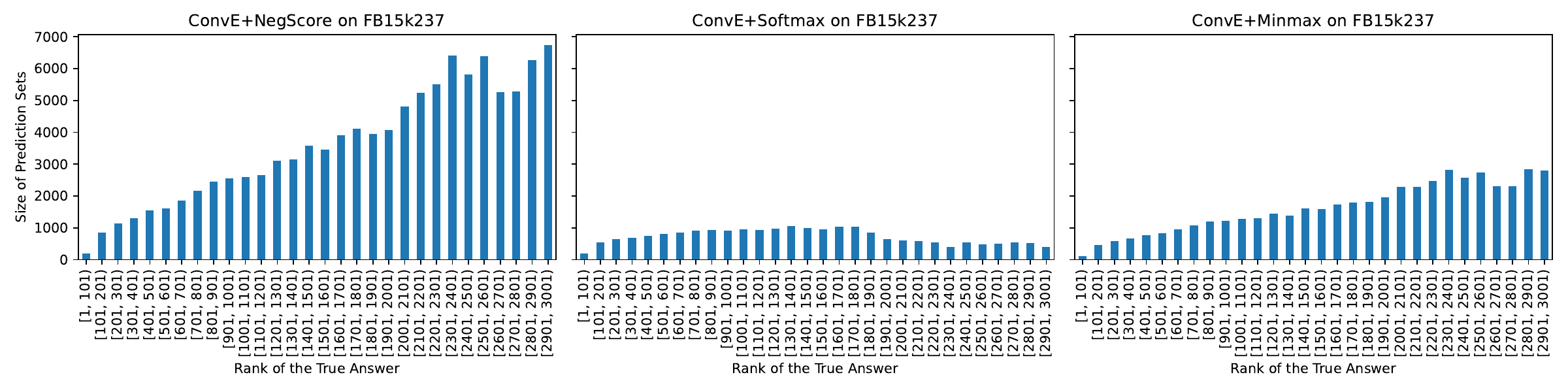}
    \caption{This figure shows the size of answer sets stratified by the difficulty level of queries. It shows the adaptiveness of different conformal predictors (built on ConvE models) on the FB15k237 dataset.}
    \label{fig:adap_ConvE_FB15k237}
\end{figure*}

In Figure \ref{fig:adap_TransE_FB15k} - \ref{fig:adap_ConvE_FB15k237}, we show the size of answer sets stratified by the difficulty level of queries for different conformal predictors across six representative KGE models and four benchmark datasets.

\section{AI Assistants In Writing}
We use ChatGPT \cite{openai2024chatgpt} to enhance our writing skills, abstaining from its use in research and coding endeavors.

\end{document}